\documentclass[sigconf]{acmart}

\PassOptionsToPackage{export}{adjustbox}
\PassOptionsToPackage{hyphens}{url}

\usepackage[normalem]{ulem}
\usepackage{adjustbox}
\usepackage{lipsum, booktabs, wrapfig}
\usepackage{amsmath}
\usepackage{graphicx}
\usepackage{mathtools}
\usepackage{soul}
\usepackage{url}
\usepackage{hyperref}
\usepackage{flushend}
\usepackage{caption}
\usepackage{subcaption}
\usepackage{multirow}

\usepackage{pgfplotstable}
\usepackage{filecontents}

\usepackage{url, hyperref}                  % hyperlink
\pgfplotsset{
    default/.style={smooth, mark options={fill=white, solid}, thick, mark size=1.50pt},
    plotStyleYellow/.style={default, Y},
	plotStyleBlack/.style={default, black},
	plotStyleEmph/.style={default, black},
	plotStyleBlue/.style={default, B},
	plotStyleRed/.style={default, R},
	plotStyleGray/.style={default, GRAY!80},
}

\pgfplotscreateplotcyclelist{myplotcyclelist}{%
solid, every mark/.append style={solid, fill=white}, mark=*\\%
solid, every mark/.append style={solid, fill=white},mark=otimes*\\%
densely dotted, every mark/.append style={solid, fill=white}, mark=otimes*\\%
loosely dotted, every mark/.append style={solid, fill=white}, mark=triangle*\\%
dashdotdotted, every mark/.append style={solid},mark=star\\%
densely dashdotted,every mark/.append style={solid, fill=white},mark=diamond*\\%
}

\usepackage{xcolor}               % graphics
    \definecolor{B}    {HTML}{2b66d3}
    \definecolor{B2}   {HTML}{003399}
    \definecolor{R}    {HTML}{c9171e}
    \definecolor{Y}    {HTML}{f1c40f}
    \definecolor{G}    {HTML}{009a00}
    \definecolor{GRAY} {HTML}{808080}
    \definecolor{MAUVE}{HTML}{9400D1}

\newcommand{\SEC}{\S}

\usepackage{colortbl}
\usepackage{tikz}
\newcommand*\Circled[1]{% require `tikz`
	\tikz[baseline=(char.base)]{\node[
        shape=circle, draw=none,  thick, 
        fill=gray!40,inner sep=0.6pt] (char) 
    {\textcolor{black}{\sffamily#1}}; 
}}

\copyrightyear{2021}
\acmYear{2021}
\setcopyright{rightsretained}
\acmConference[ICS '21]{2021 International Conference on
Supercomputing}{June 14--17, 2021}{Virtual Event, USA}
\acmBooktitle{2021 International Conference on Supercomputing (ICS
'21), June 14--17, 2021, Virtual Event,
USA}\acmDOI{10.1145/3447818.3459988}
\acmISBN{978-1-4503-8335-6/21/06}

\begin{document}

\title{ClickTrain: Efficient and Accurate End-to-End Deep Learning Training via Fine-Grained Architecture-Preserving Pruning}

\newcommand{\AFFIL}[4]{%
    \affiliation{%
        \institution{\small #1}
        \city{#2}\state{#3}\country{#4}
    }
    }

\author{Chengming Zhang}{\AFFIL{Washington State University}{Pullman}{WA}{USA}}
\email{chengming.zhang@wsu.edu}

\author{Geng Yuan}{\AFFIL{Northeastern University}{Boston}{MA}{USA}}
\email{yuan.geng@northeastern.edu}

\author{Wei Niu}{\AFFIL{College of William and Mary}{Williamsburg}{VA}{USA}}
\email{wniu@email.wm.edu}

\author{Jiannan Tian}{\AFFIL{Washington State University}{Pullman}{WA}{USA}}
\email{jiannan.tian@wsu.edu}

\author{Sian Jin}{\AFFIL{Washington State University}{Pullman}{WA}{USA}}
\email{sian.jin@wsu.edu}

\author{Donglin Zhuang}{\AFFIL{University of Sydney}{Sydney}{NSW}{Australia}}
\email{dzhu9887@sydney.edu.au}

\author{Zhe Jiang}{\AFFIL{University of Alabama}{Tuscaloosa}{AL}{USA}}
\email{zjiang@cs.ua.edu}

\author{Yanzhi Wang}{\AFFIL{Northeastern University}{Boston}{MA}{USA}}
\email{yanz.wang@northeastern.edu}

\author{Bin Ren }{\AFFIL{College of William and Mary}{Williamsburg}{VA}{USA}}
\email{bren@cs.wm.edu}

\author{Shuaiwen Leon Song}{\AFFIL{University of Sydney}{Sydney}{NSW}{Australia}}
\email{shuaiwen.song@sydney.edu.au}

\author{Dingwen Tao}{\AFFIL{Washington State University}{Pullman}{WA}{USA}}
\authornote{Corresponding author: Dingwen Tao, School of Electrical Engineering \& Computer Science, Washington State University, Pullman, WA 99163, USA.}
\email{dingwen.tao@wsu.edu}

\begin{abstract}
Convolutional neural networks (CNNs) are becoming increasingly deeper, wider, and non-linear because of the growing demand on prediction accuracy and analysis quality. The wide and deep CNNs, however, require a large amount of computing resources and processing time. Many previous works have studied model pruning to improve inference performance, but little work has been done for effectively reducing training cost. In this paper, we propose \textsc{ClickTrain}: an efficient and accurate end-to-end training and pruning framework for CNNs. Different from the existing pruning-during-training work, \textsc{ClickTrain} provides higher model accuracy and compression ratio via fine-grained architecture-preserving pruning. 
By leveraging pattern-based pruning with our proposed novel accurate weight importance estimation, dynamic pattern generation and selection, and compiler-assisted computation optimizations, \textsc{ClickTrain} generates highly accurate and fast pruned CNN models for direct deployment without any extra time overhead, compared with the baseline training.
\textsc{ClickTrain} also reduces the end-to-end time cost of the pruning-after-training method by up to 2.3$\times$ with comparable accuracy and compression ratio. 
Moreover, compared with the state-of-the-art pruning-during-training approach, \textsc{ClickTrain} provides significant improvements both accuracy and compression ratio on the tested CNN models and datasets, under similar limited training time. 
\end{abstract}




\clearpage
\maketitle
\thispagestyle{empty}
\pagestyle{empty}

\setlength{\textfloatsep}{6pt}

\section{Introduction}
\label{sec:intro}
Deep neural networks (DNNs) such as CNNs have rapidly evolved to the state-of-the-art technique for many artificial intelligence (AI) tasks in various science and technology areas, such as image and vision recognition \cite{simonyan2014very}, recommender systems~\cite{wang2015collaborative}, natural language processing~\cite{collobert2008unified}.
DNNs contain millions of parameters in an unparalleled representation, which is efficient for modeling complexity nonlinearities. 
Many works \cite{krizhevsky2012imagenet,szegedy2015going,he2016deep} have suggested that using either deeper or wider DNNs is an effective way to improve analysis quality, and in fact many recent DNNs have gone significantly deeper and/or wider \cite{wang2018superneurons,jin2019deepsz}. 
For instance, OpenAI recently published their new DNN-based NLP model GPT-3~\cite{brown2020language} with 175 billion parameters, which is the largest NLP model that is ever trained. Compared with its predecessor GPT-2, GPT-3 expands the capacity by three orders of magnitudes without significant modification to the model architecture, instead just adopting deeper and wider layers~\cite{brown2020language}.

The ever-increasing scale and complexity of the networks with large-scale training datasets such as ImageNet-2012 \cite{ILSVRC} are bringing more and more challenges to the cost of DNN training, which requires large amounts of computations and resources such as memory, storage, and I/O~\cite{wang2018superneurons,ma2018area,zhang2020enabling,zhang2021enabling,Zhang2021LSTM}. Previous works have also tried some optimization methods to try to overcome these challenges~\cite{ding2017circnn,roy2018numa,wang2018towards,ding2018structured,li2019bstc,geng2019lp}. Moreover, designing new DNN architectures and training algorithms for various AI tasks require numerous trial-and-error and fine-tuning processes, which makes the training cost issue worse and computing resources scarce.

Model pruning is a widely used approach to reduce the number of DNN weights, which can effectively reduce the computation and storage costs and increase the inference performance, especially for resource-limited platforms, such as mobile, edge, and IoT devices \cite{cai2020yolobile,li2021real,zhao2020achieving,niu2020achieving}. 
Many model pruning works have been proposed for improving the performance and energy efficiency of DNN inference \cite{molchanov2016pruning,han2016deep,luo2017thinet,yang2017designing,yu2018nisp,yuan2019sot,ma2019resnet,ma2020tiny}. 
A typical procedure to prune a DNN model consists of \Circled{1} training a model to high accuracy, \Circled{2} pruning the well-trained model, and \Circled{3} fine-tuning the pruned model. However, this procedure (called \textit{pruning-after-training or PAT}) often requires a well-trained model and a trial-and-error process with domain expertise, which is typically very time-consuming.
For example, state-of-the-art PAT-based methods~\cite{ma2020image, ma2020blk} incorporate the alternating direction method of multipliers (ADMM) into the pruning process to achieve high compression ratio and accuracy, but they almost triple the overall training time. 

Considering the weights of DNN models are gradually sparsified during training, combining the pruning and training phases together (called \textit{pruning-during-training or PDT}) is a promising way to significantly reduce the end-to-end time cost\footnote{ End-to-end time refers to the total time from the beginning of training from scratch to the end of pruning with a ready-to-deploy model.} and conserve computing resources. 
However, only few work has investigated how to prune DNN models during training while still achieving highly accurate and fast pruned models that can be directly deployed.

Recently, a state-of-the-art work \textsc{PruneTrain} \cite{lym2019prunetrain} studied how to perform CNN pruning during training. \textsc{PruneTrain} adopts a group-lasso regularization ($\ell_1$-based regularization) \cite{meier2008group} to gradually force a group of model weights with small magnitudes to zero and periodically reconfigure the CNN architecture (e.g., reducing the number of layers) during training, leading to lower computation and higher performance. 
However, in reality, adaptively changing the original network architecture may result in severe loss of accuracy, which cannot be compensated for by performing more training batches. Thus, how to design a PDT-based method to significantly reduce the end-to-end time while still maintaining the network architecture for high accuracy remains an open question.

In this paper, we propose \textsc{ClickTrain}---a fast and accurate integrated framework for CNN training and pruning---which significantly reduces the end-to-end time.
We develop a series of algorithm-level and system-level optimizations for \textsc{ClickTrain} to achieve high computation efficiency toward highly accurate and fast pruned models for inference. 
The key insights explored for algorithm-level optimization include: \Circled{1} the position of the most important weight in each convolution kernel is relatively stable after certain training batches, and \Circled{2} the important weights tend to be adjacent to each other.
Thus, we propose a fine-grained architecture-preserving pruning approach based on pattern-based pruning (will be discussed in \SEC\ref{sec:pattern}), which can preserve the original training accuracy under a higher compression ratio.
Moreover, our optimized pattern-based pruning creates multiple opportunities for system-level optimization such as sparse convolution acceleration and communication (\textsf{Allreduce}) optimization for weight update assisted by compiler so that the time overhead introduced by our proposed regularization can be mitigated.
To the best of our knowledge, our paper is \textit{the first work to study how to design a PDT-based approach for effectively reducing the end-to-end time cost while achieving very high accuracy and compression ratio of pruned models.} The main contributions are listed below:
\begin{itemize}
    \item Instead of commonly used weight selection methods based on magnitude, we incorporate a state-of-the-art weight importance estimation approach to select the desired patterns from a generated candidate pattern pool. Moreover, we propose methods to gradually generate the candidate patterns (called \textit{dynamic pattern pool generation}) and adaptively finalize the patterns and unimportant kernels.% for each kernel and the unimportant kernels.
    \item We propose a modified group-lasso regularization to replace the expensive ADMM method for pattern-based pruning.
    \item We propose multiple system-level optimizations including fast sparse matrix format conversion, pattern-accelerated sparse convolution, pattern-based communication optimization, and compiler-assisted optimized code generation, to significantly accelerate \textsc{ClickTrain} by leveraging finalized pattern sparsity during the training.
    \item We compare \textsc{ClickTrain} with state-of-the-art PAT-/PDT-based methods. Experiments illustrate that \textsc{ClickTrain} can generate highly accurate and fast pruned models for direct deployment without any extra time overhead or even faster, compared with the baseline training.
    Meanwhile, \textsc{ClickTrain} significantly reduces the training time of PAT-based approaches by up to about 2.3$\times$ with comparable accuracy and compression ratio, and improves the accuracy and compression ratio by up to 1.8\% and 4.9$\times$ over \textsc{PruneTrain}.
\end{itemize}

The rest of the paper is organized as follows. We present background information in \SEC\ref{sec:background}. We discuss our research motivation and challenges in \SEC\ref{sec:motivation}. We describe our algorithm-level design and system-level optimizations of \textsc{ClickTrain} in \SEC\ref{sec:design} and \SEC\ref{sec:computing_optimization}, respectively. We present our evaluation results in \SEC\ref{sec:evaluation}. We discuss related work and conclude our work in \SEC\ref{sec:related} and \SEC\ref{sec:con}.

\section{Background}
\label{sec:background}

\subsection{DNN Model Pruning}
Weight pruning for DNN model compression has been well studied in recent years. Below are three main methods. 

\paragraph{\bf Non-Structured Pruning.}
The non-structured pruning methods studied in the previous works \cite{zhang2018systematic} aim to heuristically prune the redundant weights on arbitrary locations.
%as shown in Figure~\ref{fig:non-structured_structured}(a). 
This leads to irregular weight distribution and inevitably introduces extra indices to store the locations of pruned weights. Eventually, this drawback limits performance acceleration \cite{dong2020rtmobile,ma2021non}.

\paragraph{\bf Structured Pruning.}
To overcome these limitations, structured pruning has been investigated in the recent studies \cite{he2018soft,he2017channel,lym2019prunetrain,yuan2019ultra,li2020ss}. They proposed to prune the entire filters, channels to maintain the structural regularity of the weight matrices after pruning.
%as shown in Figure~\ref{fig:non-structured_structured}(b). 
By taking advantage of the regular shapes of the pruned weight matrices, structured pruning becomes more hardware-friendly and achieves much higher speedups \cite{ma2019nonstructured}. However, due to the constraints of its coarse-grained pruning, structured pruning suffers from high accuracy loss.%under a higher compression ratio.

\begin{figure}
	\centering
	\includegraphics[width=1.0\linewidth]{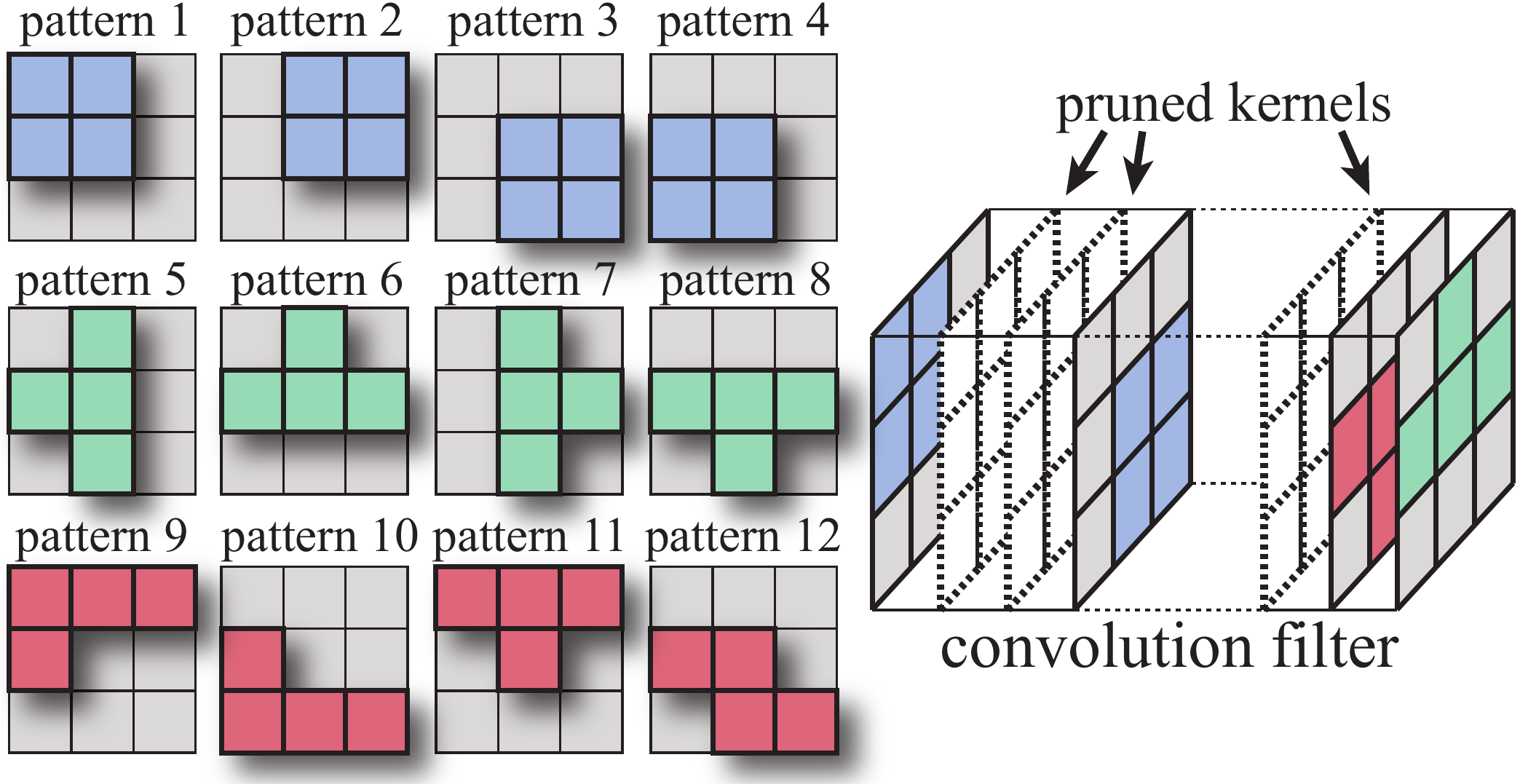}
	\vspace{-4mm}
    \caption{Pattern-based pruning with pruned kernels.}
	\label{fig:pattern_prune}
\end{figure}

\paragraph{\bf Fined-Grained Pattern-Based Pruning.}
\label{sec:pattern}
The state-of-the-art pruning work~\cite{niu2020patdnn} proposes a fine-grained pattern-based pruning scheme, which generates an intermediate sparsity type between non-structured pruning and structured pruning.
They prune a fixed number of weights in each convolution kernel (e.g., pruning 5 weights out of 9 weights in a 3$\times$3 convolution kernel), and make the remaining weights to be concentrated in a certain area to form specific kernel patterns (called \textit{pattern sparsity}), as shown in Figure~\ref{fig:pattern_prune} (left).
However, the compression ratio that is achieved by pattern sparsity is limited.
So, they further propose to exploit the inter-convolution kernel sparsity, which aims to remove some unimportant kernels (called \textit{connectivity sparsity}), as shown in Figure~\ref{fig:pattern_prune} (right). It can further enlarge the weights compression rate while reducing the convolution operations in CNNs.

The pattern-based pruning emphasizes exploiting locality in layer-wise computation, which is prevalent and widely reflected in the domains like human visual systems~\cite{93808}.
Moreover, this approach is more flexible that leads to a higher model accuracy compared to the prior coarse-grained filter/channel pruning schemes~\cite{yuan2019ultra,lym2019prunetrain}. 
Overall, a fine-grained pattern-based pruning approach, as state-of-the-art pruning scheme, considering both pattern and connectivity sparsity can leverage the advantages of non-structured and structured pruning to make the trade-off among regularity, accuracy, and compression ratio.

\subsection {Pruning-after-training (PAT) Versus Pruning-during-training (PDT)}
\textsc{GBN} \cite{you2019gate}, \textsc{GAL} \cite{lin2019towards} and \textsc{DCP} \cite{zhuang2018discrimination} are three state-of-the-art PAT-based approaches. \textsc{GBN} and \textsc{GAL} mainly focus on pruning filters. \textsc{DCP} accelerates the CNN inference via channel pruning. 
However, all of them must need well-trained models to converge to an accurate pruned model with high compression ratio, which suffers from high time cost, compared with PDT-based approaches.

\textsc{PruneTrain} \cite{lym2019prunetrain} is a state-of-the-art PDT-based approach. % to accelerate training while pruning it.
The work observed that when pruning with group-lasso regularization, once a group of model weights are penalized close to zero, their magnitudes are typically impossible to recover during the rest of the training process.
Based on this, \textsc{PruneTrain} periodically removes the small weights and change the network architecture and hence gradually reduces the training cost toward high compression ratio and accuracy.  
Moreover, \textsc{NeST} \cite{dai2019nest} and \textsc{TAS} \cite{dong2019network} are two state-of-the-art methods attempting to search the best-fit pruned network architecture during training (can be seen as PDT-based methods), but they suffer from extremely high time cost.
\section{Motivation and Challenges}
\label{sec:motivation}

\subsection{Limitations of Existing PDT-based Method}
The state-of-the-art PDT-based approach \textsc{PruneTrain} \cite{lym2019prunetrain} integrates the structured pruning with group-lasso regularization ($\ell_1$-based regularization) \cite{meier2008group} into the training phase, which is designed to prune as many channels, filters, and layers as possible to acquire a compact architecture with relatively high training performance. However, there are two major challenges to deploy \textsc{Prunetrain} for training large neural networks on a variety of architectures: inferior validation accuracy and large storage overhead.   

\begin{figure}%[ht]
	\centering
    \includegraphics[width=1.0\linewidth]{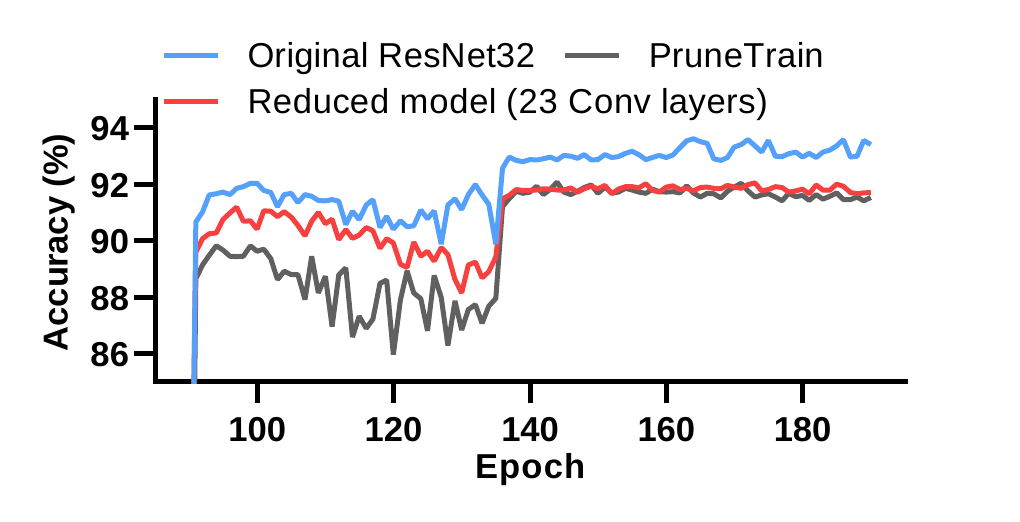}
    \vspace{-4mm}
    \caption{Achieved accuracies with three strategies: full training, \textsc{Prunetrain}, training on reduced architecture.}
	\label{fig:prunetrain-accuracy}
\end{figure}

{\bf Inferior Validation Accuracy.} 
\textsc{PruneTrain} saves the training floating-point operations (FLOPs) by drastically reconfiguring the original network architectures (e.g., reducing the depth), resulting in a notable accuracy loss for many network architectures. 
For example, \textsc{PruneTrain} can save $53\%$ training FLOPs and $66\%$ inference FLOPs (i.e., about $2.2 \times$ compression ratio) but cause $1.8\%$ accuracy drop for ResNet-32 on the CIFAR-10 dataset. Moreover, for demonstration purposes, we compare three different training strategies on ResNet-32, including (1) normal full training from scratch on the original ResNet-32 model, (2) \textsc{PruneTrain} for training the original 34-layer ResNet, and (3) training from scratch using an identical network structure as the one generated from \textsc{PruneTrain} (i.e., 23 CONV layers).
As shown in Figure \ref{fig:prunetrain-accuracy}, \textsc{PruneTrain} achieves a lower model accuracy compared to training on the reduced model, but both of them cannot reach the expected accuracy of the original ResNet-32 network, even though we train for an extra of 1,000 epochs. This experiment illustrates that the accuracy is highly relevant to network architectures. 
Moreover, studies \cite{liu2018rethinking,dong2019network,he2016deep,elsken2018neural} demonstrate that the model accuracy is highly relevant to network architecture, thus, we conclude that \textit{aggressively preserving the original architecture is critical to prevent a notable accuracy drop}.

{\bf Large Storage Overhead.} One major purpose of pruning large-scale neural networks is to facilitate their deployments in resource-constrained computing platforms which suffer from limited storage capacity and computing power. Thus, to realistically alleviate the storage and computation burden, pruning must provide a high compression ratio. However, \textsc{PruneTrain} can only provide up to $3 \times$ compression ratio (e.g., $2.2 \times$ for ResNet-32) \cite{lym2019prunetrain}, which is far below the expectation.

Therefore, these issues motivate us to develop a solution that provides \textit{three-dimensional optimizations}: high training efficiency, high compression ratio, and high model accuracy.

\subsection{Challenges of Pattern-based Pruning in Training}
Recent works \cite{ma2020image,niu2020patdnn} have applied pattern-based pruning techniques for improving inference efficiency. However, these inference-focused strategies will pose several challenges to reach our three optimization objectives.  

\begin{figure}%[ht]
	\centering
	\includegraphics[width=1.0\linewidth]{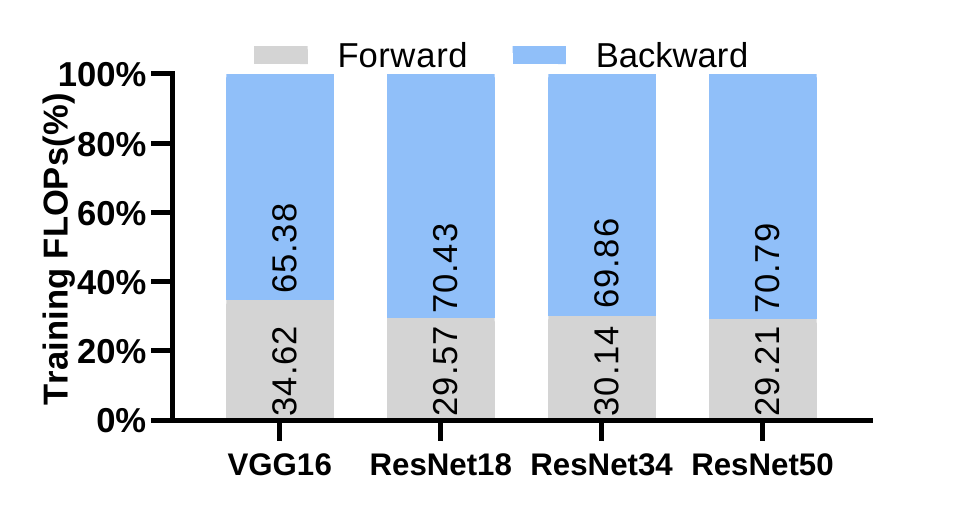}
    \vspace{-4mm}
    \caption{Percentage of floating-point operations in forward and backward phases with different CNN models.}
	\label{fig:infer_bp_percent}
\end{figure}

{\bf Issues of Existing Pattern Pruning Algorithms.}
There are three main issues of the existing pattern-based pruning algorithms \cite{niu2020patdnn, dong2020rtmobile}: 
(1) The existing algorithms select pattern for each kernel via estimating weight importance based on magnitude. However, this estimation approach requires a well-trained model, whose weights will not change dramatically, \textit{which is not true for training from scratch}.
(2) The existing algorithms predefine the candidate patterns for all kernels and statically select the best-fit pattern for each kernel. But for pruning during training, static patterns may not work properly, resulting in a significant loss of accuracy.  
(3) The existing algorithms typically with ADMM-based methods are very time-consuming and not applicable for training acceleration.

\begin{figure*}[ht]
	\centering
	\includegraphics[width=0.90\linewidth]{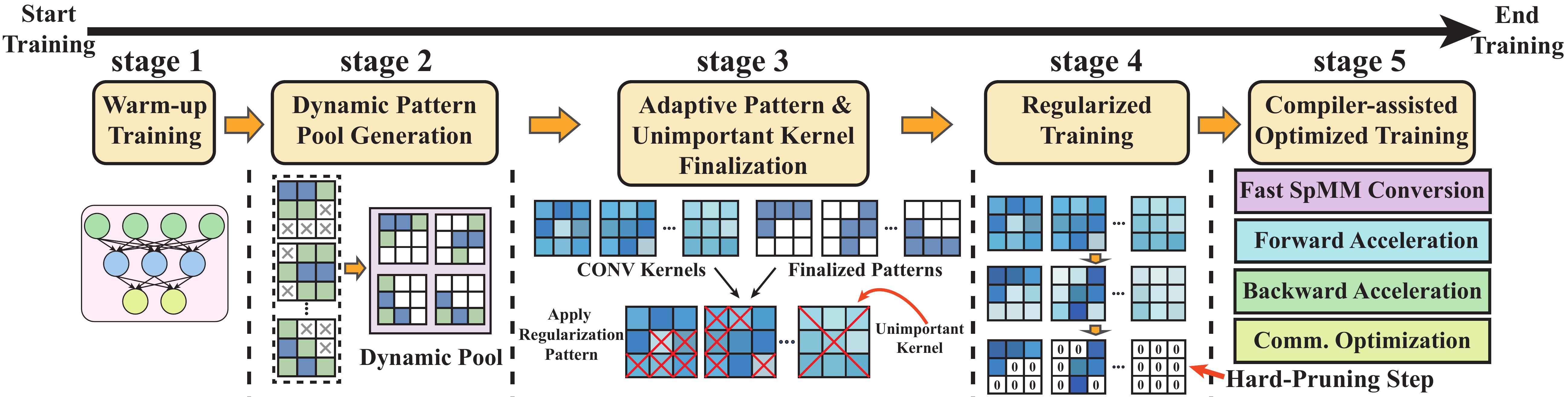}
	\vspace{-2mm}
    \caption{Overview of our proposed fast and accurate end-to-end deep learning training framework \textsc{ClickTrain}.}
	\label{fig:overall_horizontal}
\end{figure*}

{\bf Lack of Training Efficiency Optimization.}
The computation efficiency optimizations proposed by existing works \cite{niu2020patdnn,dong2020rtmobile} for pattern-based pruning cannot be directly applied to our training framework. This is mainly because of two reasons. 
On one hand, the existing pattern-based acceleration techniques \cite{dong2020rtmobile, niu2020patdnn} are designed for accelerating inference (i.e., forward phase) instead of training (including both forward and back phases). However, based on our profiling result as shown in Figure \ref{fig:infer_bp_percent}, backward phase can consume more than 70\% of the overall training FLOPs.
On the other hand, the existing pattern-based optimizations \cite{niu2020patdnn} are based on accelerating numerous convolution operations on embedded systems due to limited memory capacity. However, efficient training on advanced datacenter architectures relies on high-performance general matrix-matrix multiplication (GEMM) rather than a large number of convolutions, in order to leverage the high throughput of accelerators such as GPUs.
Thus, there is an urgent need for an effective method to take advantage of pattern-based pruning to improve the GEMM-based convolution computation efficiency.

Overall, these challenges demand a novel solution that can provide both \textit{algorithm-level} and \textit{system-level} supports for fast and accurate end-to-end training toward our three objectives. 

\section{Algorithm-Level Design of \textsc{ClickTrain}}
\label{sec:design}

In this section, we propose our novel PDT-based framework called \textsc{ClickTrain}. 
The overall framework flow is shown in Figure~\ref{fig:overall_horizontal}, which consists of five stages.
In this section, we focus on the algorithm-level design of \textsc{ClickTrain} (used in stage 2 to stage 4) mainly for quickly obtaining the model with desired pattern-based sparsity.
We first propose our pattern selection using weight importance estimation. Next, we propose our methods to dynamically build up the pattern candidates and adaptively finalize the patterns and unimportant kernels. After that, we describe our modified group-lasso regularization to accurately penalize the weights outside of the selected patterns and the unimportant kernels. 

\subsection{Pattern and Kernel Selection Criterion}
\label{sec:pattern-selection}

{\bf Weight importance estimation.}
The key consideration in the pattern-based pruning is to select the best-fit pattern for each kernel after appropriately designing the patterns.
The previous methods \cite{ma2019pconv, niu2020patdnn} determine the importance of a certain weight based on its magnitude, which requires a well-trained CNN model whose weights will not change dramatically and well distributed after pruning the redundant filters.
However, it is not feasible to accurately determine the importance of a certain weight only based on its weight magnitude during training from scratch because \textit{the weights in a not well-trained model will change greatly}, especially in the early training. Therefore, we propose to estimate the importance of patterns by further considering gradient information and to select the most important pattern for each kernel in the pruning.

For a given CNN with $L$ convolutional layers, let $W^{(\ell)}$ ($1 \leq \ell \leq L$) denote the collection of weights for all the kernels in convolutional layer $\ell$, which forms a 4-D tensor $W^{(\ell)}\in R^{F_\ell \times C_\ell \times H_\ell \times S_\ell}$, where $F_\ell$, $C_\ell$, $H_\ell$, $S_\ell$ are the dimensions for the axes of filter, channel, spatial height, spatial width, respectively. 
As suggested by \cite{molchanov2019importance}, for a weight $w_{m} \in W^{(\ell)}$, its importance can be estimated by $(g_{m}w_{m})^2$, where $g_{m} = \frac{\partial E(W, D)}{\partial w_{m}}$ is the gradient of the weight $w_{m}$. Here $E(W, D)$ is the loss function on the dataset $D$. In addition, $W$ represents the collection of all 4-D weight tensors for $L$ convolutional layers.

A pattern $p_i$ (where $p_{i} \in B$ is the $i$-th pattern from the candidate pattern pool $B = \begin{Bmatrix}p_1, p_2, \cdots,  p_{N}\end{Bmatrix}$) can be viewed as a mask to prune specific weights within a kernel. The remaining weights of the kernel form a certain pattern.
%, as shown in Figure \ref{fig:pattern_prune}. 
Thus, we can estimate the importance score of a pattern $p_i$ by combining the importance scores of all the remaining weights. 
We will discuss how to \textit{gradually generate the pattern pool} in the next section. 
The patterns corresponding to the convolutional layer $\ell$ also form a 4-D tensor $P^{(\ell)}\in R^{F_\ell \times C_\ell \times H_\ell \times S_\ell}$, 
where $P^{(\ell)}_{\smash{f_\ell,c_\ell,:,:}} \in B $. 
The importance score of $p_i$ is estimated as 
\begin{equation} \label{eq:import-est}
\textstyle t_{:,:} = G^{(\ell)}_{f_{\ell}, c_{\ell},:,:} \odot W^{(\ell)}_{f_{\ell}, c_{\ell},:,:} \odot p_{i}, \;\;
I_{p_{i}} = \sum_{h_\ell}^{H_\ell} \sum_{s_\ell}^{S_\ell} (t_{h_{\ell} \times s_{\ell}} )^2
\end{equation}
where $G^{(\ell)}$ denotes the 4-D gradient tensor corresponding to $W^{(\ell)}$ and $\odot$ is the element-wise product (a.k.a, Hadamard product). After assessing the importance of all the patterns for a kernel, we can choose the pattern with the highest estimated importance score as the best-fit pattern for this kernel.

In addition, to further enlarge the sparsity, we also need to determine the unimportant kernels and directly prune them (i.e., connectivity sparsity). We adopt the following equation to estimate the importance score of a kernel $k_{i} \in W^{(\ell)}$.
\begin{equation} \label{eq:unimport-est}
\textstyle t_{:,:} = G^{(\ell)}_{f_{\ell}, c_{\ell},:,:} \odot W^{(\ell)}_{f_{\ell}, _{\ell},:,:}, \;\;
I_{k_{i}} = \sum_{h_\ell}^{H_\ell} \sum_{s_\ell}^{S_\ell} (t_{h_{\ell} \times s_{\ell}} )^2
\end{equation}

The reason why we integrate gradient to estimate the importance is if a certain weight whose magnitude and gradient are small, it will very likely keep the small value in the following training process because the backpropagation tends not to update its value dramatically. 
Therefore, we can treat it as an unimportant weight and penalize such weights during training to push it to become smaller and smaller (less and less important).
Eventually, we can prune it with no hurt to the final accuracy.
The computation cost of Equation (\ref{eq:import-est}) and (\ref{eq:unimport-est}) are relatively low, since the gradient $G^{(\ell)}$ can be acquired in the backward-propagation stage, which can be naturally implemented in most of the deep learning frameworks \cite{abadi2016tensorflow,paszke2019pytorch}. 
Moreover, the number of candidate patterns is limited to a relatively small number (will be discussed later).

\begin{figure}%[ht]
    \centering 
    \includegraphics[width=1.0\columnwidth]{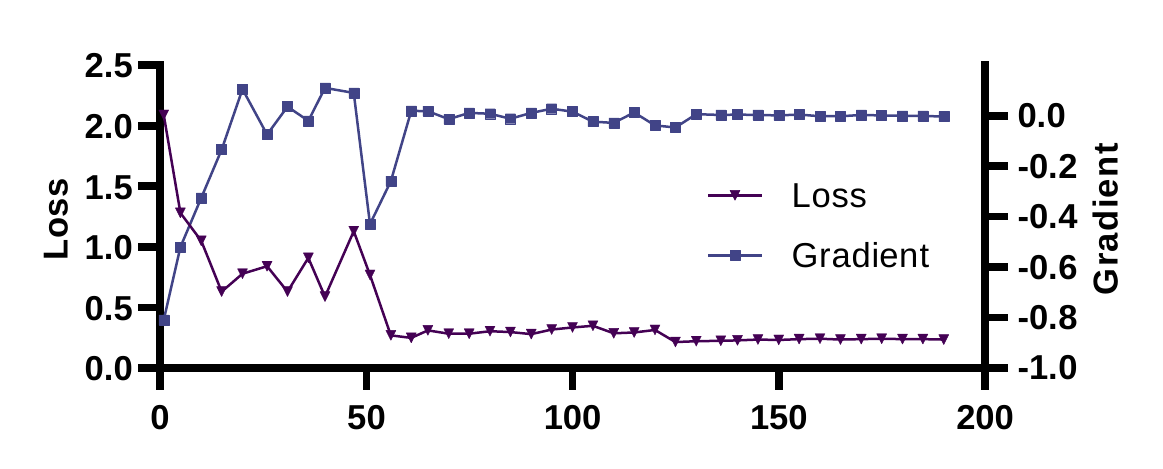}
    \vspace{-2mm}
    \caption{Loss \& gradient in training ResNet-32 on CIFAR-10.}
%    \vspace{-4mm}
    \label{fig:loss-gradient}
\end{figure}    

{\bf Pattern and kernel selection.}
The weights and gradients would change greatly during the first few epochs of training CNN.
On one hand, estimating the weight importance earlier can help us remove these unimportant weights sooner, thereby reducing the overall training time.
On the other hand, estimating the importance of weights prematurely will lead to inaccurate estimations and ultimately cause a significant accuracy drop. Later estimation does help to improve the pruning accuracy but would cost more training epochs. 
Therefore, when we should apply the formulas above to accurately assess the weight importance is a challenge to address.
After a series of empirical evaluation, we conclude that \textit{the derivative of the loss function on epochs can be used as a good indicator to solve this problem}.
In particular, when this value is less than a threshold, 
we can apply the formula above to evaluate the weight importance relatively accurately. 
For example, Figure \ref{fig:loss-gradient} shows that the loss does not change sharply after a certain threshold (50 epochs), meaning that the optimization process gradually stabilizes to start our pruning.% via the importance estimation. 

\subsection{Dynamic Pattern Pool Generation}
\label{sec:DPPG}

{\bf Limitations of static pattern pool.}
As discussed in Section~\ref{sec:pattern}, using 3$\times$3 kernel as an example, a pattern can be formed by any 4 positions inside a kernel, which will result in a total number of $(^9_4)=126$ possible different pattern types.
In general, a larger number of pattern types used in a CNN will lead to a considerable runtime overhead, offsetting the training acceleration. 
On the other hand, too few pattern types will reduce the pruning flexibility and hence accuracy degradation.
To preserve a high accuracy while not compromising computation efficiency, we need to limit the number of candidate pattern types that can be used for pruning.

Moreover, unlike pruning a well-trained CNN, pruning during training faces the challenge that the positions of importance weights are not stabilized.
It is not ideal to determine the positions of pruned weights only once and fix them in the subsequent training process.

Therefore, we propose the Dynamic Pattern Pool Generation (\textit{DPPG}) method, which selects important weights gradually to build a pattern pool with a limited number of desired pattern types.
A multi-stage procedure is proposed to build pattern pool dynamically. 
In this work we use 3$\times$3 kernels for demonstration, but our proposed method can also be applied to other kernel sizes, such as 5$\times$5, 7$\times$7.

\begin{figure}%
    \centering
    \includegraphics[width=1\columnwidth]{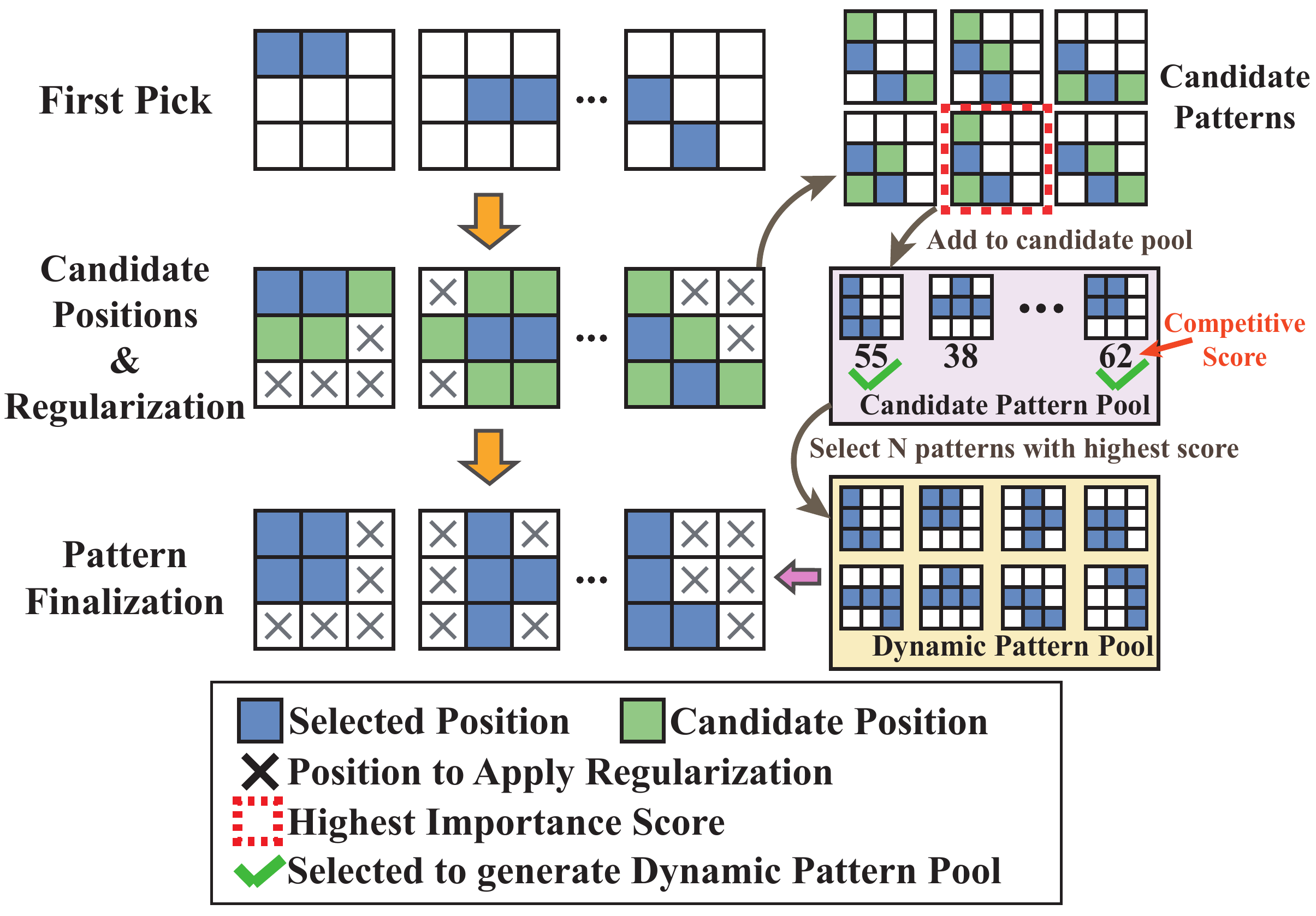}
    \vspace{-4mm}
    \caption{Our proposed DPPG method.}
    \label{fig:build_pattern}
\end{figure}

{\bf Proposed dynamic pool generation process.}
The whole process mainly consists of the following six steps. \Circled{1} We first select one weight position with the highest estimated importance score in each kernel. 
\Circled{2} We then select the second weight position with the highest importance score among the adjacent positions of the first position, including its horizontal, vertical, and diagonal adjacent position.
\Circled{3} After determining the first two positions in patterns (marked in blue in Figure \ref{fig:build_pattern}), 
we mark the adjacent positions to the first two positions as the ``candidate positions'' (marked in green).
\Circled{4} For each kernel, we pick two candidate positions together with the first two positions to form a candidate pattern. 
We evaluate the importance score (as mentioned in Section~\ref{sec:pattern-selection}) of all possible combinations (i.e., candidate patterns) to find the candidate pattern with the highest score, then add it to a global candidate pattern pool. 
Each candidate pattern in the candidate pattern pool has a competitive score that is initialized to 1. 
If the candidate pattern already exists in the candidate pattern pool, then we add ``1'' to its competitive score.
\Circled{5} We repeat the step 1$\sim$4 for a certain number of epochs (depending on the training epoch budget), and keep accumulating the competitive scores in the candidate pattern pool.
\Circled{6} Finally, we obtain the desired final pattern pool by selecting $N$ candidate patterns from candidate pattern pool with the highest competitive scores, where $N$ is the given number of total pattern types (usually is set to 8 or 12).

{\bf Detailed optimizations.}
In order to better cluster the candidate patterns in the candidate pattern pool and generate a desired pattern pool, we apply several optimizations/constraints during the dynamic pattern pool generation process.

For \Circled{2}, the selection rule of the second weight position is derived by considering two observations: 
i) the position of the most important weight (i.e., the highest-scoring weight) in each kernel is relatively stable after certain training iterations, and ii) the weight distribution in unpruned (dense) well-trained models suggest that the important weights tend to be adjacent to each other.

For \Circled{3}, since the weights on the candidate positions are usually less important than the firstly selected two weights, model has a higher tolerance for these weights not being selected optimally.
Thus, we intentionally exclude the diagonal adjacent positions to reduce the diversity of the candidate positions to cluster the candidate patterns.
Moreover, by limiting the possible candidate positions in \Circled{3}, part of the pruning positions (marked as ``$\times$'' in Figure~\ref{fig:build_pattern}) in each kernel can be determined earlier, thereby starting the regularization and fine tuning sooner, which both reduces the training time and enhances the pattern formation process.

\subsection{Adaptive Pattern and Kernel Finalization}
\label{sec:Pattern-Finalization}

After the pattern pool is generated, we can finalize the pattern for each kernel adaptively based on the pattern pool.
Due to the concern that the one-time selection method may not provide high accuracy, we propose an adaptive method to finalize the patterns using multiple training epochs.
Specifically, in each training batch, for every kernel, we calculate the importance score of each pattern in our pattern pool.
Then, we find the pattern with the highest importance score and count its number of occurrences during training. After a number of epochs, the final pattern for each kernel will be selected as the most frequent one of those highest-scoring patterns. 

\begin{figure*}[ht]
	\centering
	\includegraphics[width=\linewidth]{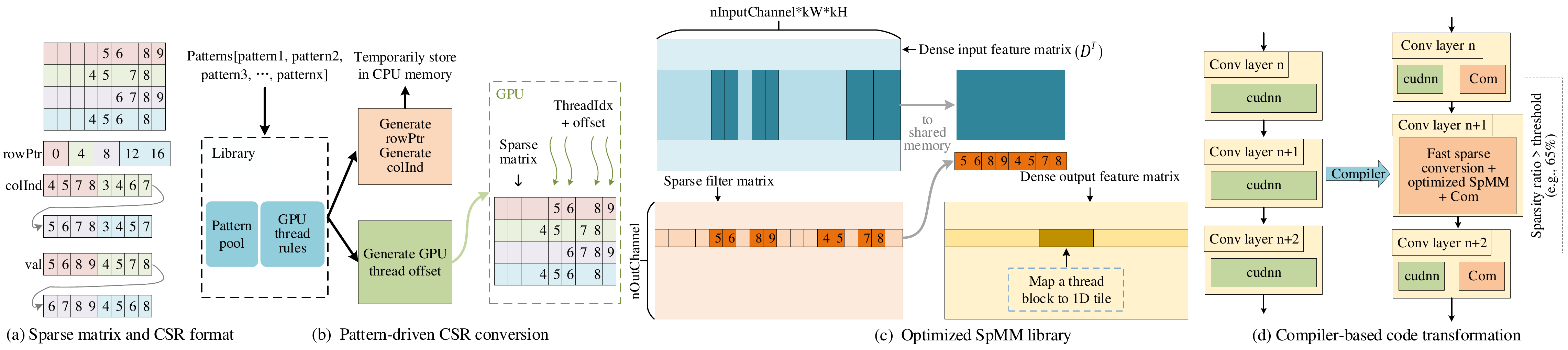}
	\vspace{-6mm}
    \caption{Our proposed compiler-assisted pattern-accelerated SpMM for sparse convolution.}
	\label{fig:CSR}
\end{figure*}

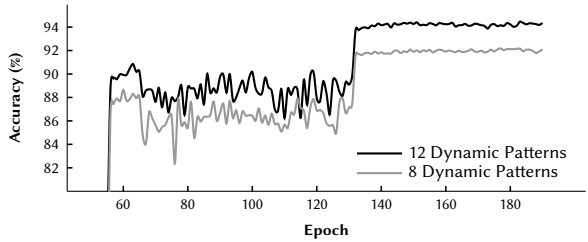
\begin{figure}[ht]
    \centering\sffamily
    \begin{tikzpicture}
\begin{axis}[
    width=1.0\linewidth,
	height=1.6in,
	label style={font=\footnotesize},
	cycle list name=myplotcyclelist,
	legend cell align={left},
	legend columns=1, 
	legend style={
		at={(0.99,0.15)},anchor=east,
		font=\footnotesize\sffamily,
		/tikz/column 4/.style={column sep=5pt,},
		row sep=-2.5pt,
		draw=none, draw opacity=1,
		fill=none, fill opacity=1,
		text opacity=1
	},
	tick pos = left,
	tick label style={font=\scriptsize\sffamily},
    major tick length=2pt,
    every tick/.style={black, semithick,},
	    xtick={60, 80, 100, 120, 140, 160, 180},
	    xticklabels={60, 80, 100, 120, 140, 160, 180},
	    ytick={82, 84, 86, 88, 90, 92, 94},
	    yticklabels={82, 84, 86, 88, 90, 92, 94},
        restrict expr to domain={y}{70:96},
        ymin=80,
	axis x line*=bottom,
	axis y line*=left,
	xlabel={Epoch},
	ylabel={Accuracy (\%)},
	x label style={font={\bfseries\scriptsize}, at={(0.5,0.08)}},
	y label style={font={\bfseries\scriptsize}, at={(0.08,0.5)}},
]

\pgfplotstableread[col sep=space, header=true]{timeline.tsv}\data;

\addplot+ [plotStyleEmph, mark=none] table [y index=2,  x index=0,  col sep=tab] {\data};
\addplot+ [plotStyleGray, axis on top=true, mark=none] table [y index=1,  x index=0,  col sep=tab] {\data};

\legend{
    12 Dynamic Patterns,
    8 Dynamic Patterns
    }
\end{axis}
\end{tikzpicture}
    \vspace{-2mm}
    \caption{Accuracy comparison between 8 and 12 dynamic patterns.}
    \label{fig:accuracy_pattern_selection}
\end{figure}

Similarly, we also use our importance estimation method to calculate the importance score for each kernel and adaptively select a user-set number of unimportant kernels for each layer. 
Note that the loss is not always decreasing during training. Thus, we set up a training loss margin $\delta$ as a hyperparameter to avoid selecting patterns in the training batches where the loss is obviously increasing.
For example, if the loss in the previous batch divided by the value in the current batch is smaller than $0.0018$ (default $\delta$), we will not count the highest-scoring patterns into the number of occurrences.
In addition, Figure \ref{fig:accuracy_pattern_selection} illustrates that 12 dynamic patterns achieves $2\%$ higher than the one-time solution on CIFAR-10 since kernels have greater probability of selecting the appropriate pattern.

\subsection{Modified Group Lasso Regularization}

$\ell_1$-based regularization or group-lasso regularization is usually added to the loss function to penalize all important or unimportant weights along desired dimensions (such as filter, channel) over the entire layers of CNNs.
However, we have noted that using group-lasso regularization to penalize all the weights of filters or channels can lead to a \textit{severely impaired pruned model}. Thus, we propose a modified group-lasso regularization to more accurately penalize the weights. 
Generally speaking, unimportant weights should be punished more heavily than important weights, and important weights should remain the same because they typically play a key role in generating stronger activation to make more confident decisions.
In particular, after selecting the best-fit pattern for each kernel and identifying the unimportant kernels, we only penalize the unimportant kernels and the weights outside of the selected patterns, since we desire to reduce the absolute values of these weights and kernels as the training progresses. 
We note that the contributions of these weights/kernels to the final accuracy are negligible, so even if we directly remove (i.e., hard prune) these weights/kernels, the model accuracy would not decrease obviously.

Let $I^{(\ell)}$ be a 4-D importance tensor (i.e., a tensor full of importance scores) of a 4-D weight tensor $W^{(\ell)}$, where $I^{(\ell)}$ is the same shape as $W^{(\ell)}$.
Assume the $c_\ell$-th kernel of the $f_\ell$-th filter in the convolutional layer $\ell$ is unimportant, 
we set $I^{(\ell)}_{\smash{f_\ell,\ c_\ell,:,:}}$ to be 0. 
Our proposed approach for penalizing unimportant weights and kernels with modified group-lasso regularization can be formulated as:

\begin{equation} \label{eq:import-est0}
Z^{(\ell)} = W^{(\ell)} \odot \left(\neg P^{(\ell)} \right), \;\;
U^{(\ell)} = W^{(\ell)} \odot \left(\neg I^{\hspace{.15em}(\ell)} \right) 
\end{equation}
\vspace{-2mm}
\begin{equation} \label{eq:import-est1}
\begin{split}
\scriptstyle E(W, D) \scriptstyle &= E(W, D) + \lambda_P \sum_{l=1}^{L} \left(\sum_{f_l=1}^{F_l} \sum_{k_l=1}^{K_l} \left \|  Z^{(l)}_{f_l,k_l,:,:} \right \|_g\right) \scriptstyle \\ &+ \lambda_{\mathclap{\ \ I}\phantom{P}} \sum_{l=1}^{L} \left(\sum_{f_l=1}^{F_l} \sum_{k_l=1}^{K_l} \left \| U^{(l)}_{f_l,k_l,:,:} \right \|_g \right)
\end{split}
\end{equation}

\noindent where $\scriptstyle\left \| w^{(g)} \right \|_g$ = $\scriptstyle\sqrt{\sum_{i=1}^{|w^{(g)}|} \left (w_i ^{(g)} \right )^2}$, $|w^{(g)}|$ is the number of weights in $ w^{(g)}$, $\neg$ is the element-wise inversion operator (i.e., 0 $\rightarrow$ 1 or 1 $\rightarrow$ 0), and $\lambda$ is the coefficients for the group-lasso regularization.
Here the inverse tensors $\neg P^{(\ell)}$ and $\neg I^{(\ell)}$ can be considered as masks, which can prevent from unnecessarily penalizing the important weights. Moreover, these tensor operations can be accelerated by many popular GPU-based deep learning frameworks such as TensorFlow and PyTorch \cite{abadi2016tensorflow,paszke2019pytorch} to speedup the group lasso.
\section{System-Level Optimizations of ClickTrain}
\label{sec:computing_optimization}
In this section, we discuss our proposed system-level optimizations (used in stage 5) for improving training computation efficiency.

\subsection{Pattern-driven Fast Sparse Matrix Conversion}

Kernels become much more sparse than origins after pruning. According to prior studies \cite{choy20194d, yao2019balanced}, a highly optimized sparse matrix-matrix multiplication (SpMM) leveraging pruning sparsity can outperform state-of-the-art GPU GEMM libraries such as cuBLAS \cite{cublas} for convolution computation. 
SpMM typically requires converting dense input matrix to a sparse format such as Compressed Sparse Row (CSR) because CSR dominates a continuous storage space, which is beneficial to data locality and computation efficiency.

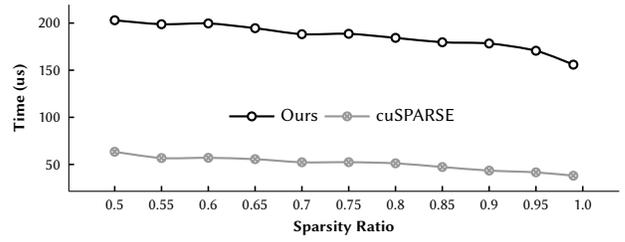
\begin{figure}%[ht]
	\centering\sffamily
    \begin{tikzpicture}[baseline]
\begin{axis}[
    width=1.05\linewidth,
	height=1.6in,
	label style={font=\footnotesize},
	cycle list name=myplotcyclelist,
	legend cell align={left},
	legend columns=2, 
	legend style={
		at={(0.5,0.3)},anchor=south,
		font=\footnotesize\sffamily,
		/tikz/column 4/.style={column sep=5pt,},
		row sep=-2.5pt,
		draw=none, draw opacity=1,
		fill=none, fill opacity=1,
		text opacity=1
	},
	tick pos = left, % equivalent to "xtick pos=left, ytick pos=left"
	tick label style={font=\scriptsize\sffamily},
    major tick length=2pt,
    every tick/.style={black, semithick,},
	    xtick={0.5, 0.55, 0.6, 0.65, 0.7, 0.75, 0.8, 0.85, 0.9, 0.95, 1.0},
	    xticklabels={0.5, 0.55, 0.6, 0.65, 0.7, 0.75, 0.8, 0.85, 0.9, 0.95, 1.0},
		ytick={50,100,150,200},
		yticklabels={50,100,150,200},
% 	hide axis,
	axis x line*=bottom,
	axis y line*=left,
	xlabel={Sparsity Ratio},
	ylabel={Time (us)},
	x label style={font={\bfseries\scriptsize}, at={(0.5,0.1)}},
	y label style={font={\bfseries\scriptsize}, at={(0.08,0.5)}},
]

\pgfplotstableread[col sep=space, header=true]{csr_conversion_time.tsv}\data;

\addplot+ [plotStyleEmph] table [y index=2,  x index=0,  col sep=tab] {\data}; %% ours
\addplot+ [plotStyleGray] table [y index=1,  x index=0,  col sep=tab] {\data};  %% comparison

\legend{Ours, cuSPARSE}
\end{axis}
\end{tikzpicture}
    \vspace{-2mm}
    \caption{CSR conversion time.}
%    \vspace{-4mm}
    \label{fig:convert_time}
\end{figure}

However, converting a dense matrix to its CSR format (as shown in Figure \ref{fig:CSR} (a)) usually introduces an obvious time overhead if directly calling \textsf{dense2csr()} from cuSPARSE library \cite{cusparse}.
For example, as shown in Figure~\ref{fig:convert_time}, \textsf{dense2csr()} costs about 200 us when converting a 256$\times$1152 dense matrix to its CSR format (generated by the convolution operation on 128$\times$224$\times$224 and 256$\times$128$\times$3$\times$3 tensors), whereas the time of the corresponding SpMM of 256$\times$1152 (sparse) multiplying 1152$\times$50176 (dense) is only about 3000 us.
We note that after the patterns have been finalized, the positions of un-pruned (non-zero) weights will not change. 
Thus, we can directly generate \verb+rowPtr+, \verb+colInd+, and GPU thread \verb+offset+ arrays (as shown in Figure \ref{fig:CSR} (b)) during stage 3 (as shown in Figure \ref{fig:overall_horizontal}) and store them for the following SpMM-based convolution operations. 
Note that this indices generation only introduces a negligible time overhead, since we only need to generate the fixed arrays for once and keep using them in the next stages. 

Moreover, since \textsf{dense2csr()} does not provide any interface for pre-defined nonzero positions (rowPtr and colInd), we propose a fast dense-to-sparse matrix conversion routine with pre-defined nonzero positions, whose interface as follow.

% \begin{center}
% \adjustbox{max width=\linewidth}
{
\vspace{1mm}
\ttfamily
template <typename T>
\textsf{{Convert2CSR}}(%
int* rowPtr, 
int* colInd, 
T* sparseFilters, 
T* val)
\vspace{1mm}
}

% \end{center}
\noindent Note that in order to maintain a minimal modification to the existing popular deep learning frameworks such as PyTorch which uses dense matrices for most computations such as \texttt{autograd} (automatic differentiation), we apply our fast dense-to-sparse matrix conversion before each SpMM (for all filters in one layer) rather than changing all computations to be based on sparse matrices.%in this work.

For demonstration purpose, we evaluate our proposed fast matrix conversion on the 256$\times$1152 and 1152$\times$50176 matrices using an NVIDIA RTX 5000 GPU and compare it with cuSPARSE. Figure~\ref{fig:convert_time} illustrates that our conversion implementation is 4$\times$ faster than cuSPARSE's \textsf{dense2csr()}.

\subsection{Pattern-Accelerated SpMM for Sparse Convolution}
Ideally, we can save the floating-point operations if not involving the pruned weights (zeros) in the convolution operation.
However, even though pattern sparsity is more regular than random sparsity (unstructured pruning), existing hardware such as GPUs cannot utilize the pattern sparsity to accelerate either forward or backward phase.
We observe that our pattern sparsity exhibits three key characteristics: 
\Circled{1} The types of sparsity is relatively limited, such as 4, 8, or 12. 
\Circled{2} The non-zero (un-pruned) weights inside a kernel are more likely close to each other. 
\Circled{3} Each kernel has the same number of non-zero weights. 
Therefore, considering that SpMM can decrease the computational cost for sparse convolution operations by reducing the number of multiplications and additions, we propose to design a new GPU SpMM library for sparse convolution to make full use of the above important characteristics and use CUDAAdvisor~\cite{shen2018cudaadvisor} to guide CUDA code optimization.

{\bf Pattern-Accelerated SpMM.}
We first describe our design of pattern-accelerated GPU SpMM library that exploits the pattern sparsity in both forward and backward phases. 
Specifically, converted sparse matrices after pruning lose the regular structure of dense matrices, which results in irregular memory accesses in SpMM. 
Moreover, concurrently executing massive threads on GPU makes the random memory access issue worse. 
Thus, improper handling of random memory accesses from massive parallel threads would stall the thread execution and decrease the performance significantly. 
In order to overcome the challenge of random memory accesses, we propose to take advantage of shared memory on GPU architectures to support concurrent random accesses. 
Specifically, we first load a tile of data from input feature matrix (dense) and filter matrix (sparse) to shared memory, as shown in Figure~\ref{fig:CSR} (c). Then, we use the loaded data (in shared memory) to calculate the corresponding tile of output feature matrix (dense). 
Inspired by existing works \cite{Guyue2020general,Trevor2020sparse}, load imbalance may severely hurt the performance on the GPU, while we solve this issue through an algorithm-hardware co-design. 
On the algorithm side, we limit all the filters in the same layer have the same number of un-pruned (non-zero) weights in our pattern-based pruning.
On the hardware side, we can further improve the performance by using the vectorized load and store instructions in CUDA architectures \cite{ProTip} because each row of sparse filter matrix has the same length (i.e., non-zero weights). 
In addition, since the sparse matrices generated by convolution operations typically have relative long rows, we adopt 1D tiling strategy and map each thread block to a 1D %row tile of the output matrix. 

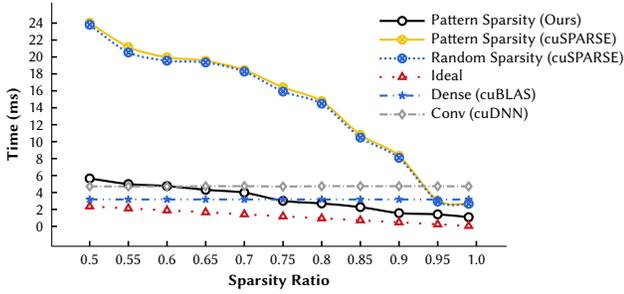
\begin{figure}%[ht]
	\centering\sffamily
    \begin{tikzpicture}
\begin{axis}[
    width=.9\linewidth,
	height=1.9in,
	label style={font=\footnotesize},
	cycle list name=myplotcyclelist,
	legend cell align={left},
	legend columns=1, 
	legend style={
		font=\scriptsize\sffamily,
		/tikz/column 4/.style={column sep=5pt,},
		row sep=-2.5pt,
		draw=none, draw opacity=1,
		fill=none, fill opacity=1,
		text opacity=1,
        at={(.7,1)},
        anchor=north west
	},
	tick pos = left,
	tick label style={font=\scriptsize\sffamily},
    major tick length=2pt,
    every tick/.style={black, semithick,},
	    xtick={0.5, 0.55, 0.6, 0.65, 0.7, 0.75, 0.8, 0.85, 0.9, 0.95, 1.0},
	    xticklabels={0.5, 0.55, 0.6, 0.65, 0.7, 0.75, 0.8, 0.85, 0.9, 0.95, 1.0},
	    ytick={0,2,4,6,8,10,12,14,16,18,20,22,24},
	    yticklabels={0,2,4,6,8,10,12,14,16,18,20,22,24},
	axis x line*=bottom,
	axis y line*=left,
	xlabel={Sparsity Ratio},
	ylabel={Time (ms)},
	x label style={font={\bfseries\scriptsize}, at={(0.5,0.08)}},
	y label style={font={\bfseries\scriptsize}, at={(0.12,0.5)}},
]

\pgfplotstableread[col sep=space, header=true]{more_conversion_time.tsv}\data;

\addplot+ [plotStyleEmph, axis on top=true] table [y index=1,  x index=0,  col sep=tab] {\data};
\addplot+ [plotStyleYellow] table [y index=2,  x index=0,  col sep=tab] {\data};
\addplot+ [plotStyleBlue] table [y index=3,  x index=0,  col sep=tab] {\data};
\addplot+ [plotStyleRed] table [y index=4,  x index=0,  col sep=tab] {\data};
\addplot+ [plotStyleBlue] table [y index=5,  x index=0,  col sep=tab] {\data};
\addplot+ [plotStyleGray] table [y index=6,  x index=0,  col sep=tab] {\data};

\legend{
    {Pattern Sparsity (Ours)},
    {Pattern Sparsity (cuSPARSE)},
    {Random Sparsity (cuSPARSE)},
    Ideal,
    Dense (cuBLAS),
    Conv (cuDNN)
    };
\end{axis}
\end{tikzpicture}
    \vspace{-4mm}
    \caption{Convolution time with different methods.}
%    \vspace{-4mm}
	\label{fig:sparse_infer}
\end{figure}

For demonstration purpose, we also evaluate our optimized SpMM on those 256$\times$1152 and 1152$\times$50176 matrices using an NVIDIA RTX 5000 GPU and compare it with state-of-the-art dense or sparse GEMM libraries. Figure~\ref{fig:sparse_infer} illustrates that our optimized SpMM library can achieve a speedup of 4.5$\times$ over cuSPARSE. Moreover, our implementation is faster than cuBLAS's GEMM and cuDNN's convolution when the sparsity ratio is higher than 65\%.

{\bf Forward Phase Acceleration.}
To achieve a final high accuracy, the filters in different layers may have completely different compression/sparsity ratio. We note that for relatively low sparsity ratio, our optimized SpMM does not gain a performance improvement compared to its dense counterpart (will be showed later). Thus, we set a threshold of sparsity ratio for each layer, and only call our optimized SpMM for the layer where its sparsity ratio is higher than the threshold in the forward phase.  
%and we will call our customize GPU library only after the sparse ratio of this layer is higher than those sparse ratio thresholds. 
Based on experiment, we set 65\% as the default threshold for all layers without loss of generality.
%we determine those thresholds based on theoretical and experimental analysis.
%We will show our result that our optimized forward phase can achieve an average speedup of 1.3$\times$.

{\bf Backward Phase Acceleration.}
%we attain full utilization of the GPU hardware, where a ready warp can be run on every cycle, all computational units are doing useful work on every cycle, and all memory accesses are coalesced. Our principles for reaching this goal are (1) effective latency hiding through a combination of thread- and instruction-level parallelism (TLP and ILP) and (2) efficient load-balancing. the load-balancing include: Load imbalance across warps. Some warps may be assigned less work than others, which may lead to these less-loaded computation units being idle while the more loaded ones continue to do useful work. Load imbalance within a warp: (a) Some warps may not have enough work to occupy all 32 threads in the warp. (b)Some warps may assign different tasks to different threads. Convolution is the predominant operation in CNNs.
The output $z^\ell$ (before activate function $\sigma$) of the convolutional layer $\ell$ in the CNNs' forward-propagation is obtained by $z^\ell = a^{\ell-1} \circledast W^\ell + b^\ell$, where $a^{\ell-1}$ is the activations at layer $\ell-1$, $W^\ell$ and $b^\ell$ denote weights and biases at layer $\ell$, respectively, and $\circledast$ is the convolution operation.
In backpropagation, the layer $\ell$ first receives $\delta^{\ell}$ from the layer $\ell+1$, and then $\delta^{\ell}$ is propagated back based on
$
\delta^{\ell} =  {\textstyle \frac{\partial E(W, D)}{\partial z^{\ell}} },
\delta^{\ell-1} = \delta^{\ell} \circledast \text{\sffamily rot180}(W^\ell) \odot \sigma^{\prime}(z^{\ell-1}).
$
We can calculate the gradient of the layer $\ell$ after obtaining $\delta^{\ell}$ and
$\frac{\partial E(W, D)}{\partial W^{\ell}} = a^{\ell-1} \circledast \delta^{\ell}$.
We can observe from the above equation that the difference between forward and backward phases is that the forward phase uses feature map $a^{\ell-1}$ as input, whereas the backward phase uses $\delta^{\ell+1}$ as input. Also, the sparse weight matrix $W^{\ell}$ is involved in the backpropagation, so we adopt the similar strategy as forward phase to handle $W^{\ell}$. 
%We will show our experimental result that our optimized backward phase can achieve an average speedup of 1.2$\times$. 

\subsection{Communication Optimization}
Distributed training has been widely used for larges-scale DNN training. However, gradients must be synchronized among different computing nodes for each training batch, such communication (i.e., \textsf{Allreduce}) overhead is not negligible and will be scaled up as the number of computing nodes increases.
% Thanks to the fixed positions of non-zero (un-pruned) weights after finalizing the patterns, we only need to pass the gradients corresponding to those positions and update their weights after the hard pruning.
Note that since all the weights to be pruned remain zeros after the regularized training stage, we do not need to send the corresponding gradients to other computing nodes in the rest of the training process.
%\textcolor{red}{Note that although the positions of pruned weights have been determined when the patterns are selected, these weights are actually set to zeros after regularized pruning stage.}
Moreover, high sparsity ratio of our pattern-based pruning  provides a great opportunity to significantly reduce the communication overhead. 

\subsection{Compiler-Assisted Optimized Code Generation}

After implementing our optimized libraries for sparse matrix conversion, SpMM, and \textsf{Allreduce}, we proposes a compiler-assisted method to generate the optimized code for efficient training (in stage 5).
Specifically, after sparsity ratios have been determined (after stage 3), the compiler decides whether using original sparse convolutions or our optimized SpMM computation for each layer in the computational graph based on its sparsity ratio; and accordingly generates the code with the better operator in the training framework such as PyTorch (as shown in Figure \ref{fig:CSR} (d)). 
Moreover, the compiler transforms all \textsf{Allreduce} communications (in stage 5) in the computational graph into our optimized pattern-based communications.
Note that there is an alternative solution that relies on compiler to generate the shared library calling sparse convolutions or SpMM computation dynamically for each layer. However, this solution introduces ``if-else'' overhead for each layer in every training batch, which is much higher than the former solution. 

%To eliminate redundant computation between pruned weight and input matrix, \textsc{ClickTrain} proposes on a compiler-assisted framework to achieve efficient training. This compiler framework generates optimized code according to the network architecture, such as feature size, input channel, and output channel. Besides conventional optimization in compiler system, e.g. tiling, blocking, loop permutation optimizations, our framework also supports filter load redundant elimination and reorder to solve the key challenges---sparse data reuse and load imbalance. These challenges are well solved in dense matrix computation, while they are still challenging in sparse matrix computation. After code generation, the generated shared library will be adopted in host framework, e.g. Pytorch or TensorFlow. We then dynamically choose a runtime scheduling policy to determine whether using sparse CONV or SPMM computation in the shared library.
\section{Experimental Evaluation}
\label{sec:evaluation}
In this section, we first evaluate our proposed \textsc{ClickTrain} on different CNNs and datasets and show its accuracy and compression ratio and compare it with several state-of-the-art PDT-/PAT-based frameworks. We then evaluate our proposed optimizations and overall training efficiency on a single GPU and multiple GPUs. 

\subsection{Experimental Setup}
We conduct our experimental evaluation using the Frontera supercomputer \cite{stanzione2020frontera} at TACC, of which each GPU node is equipped with two Intel E5-2620 v4 CPUs and four NVIDIA Quadro RTX 5000 GPUs \cite{rtx5000}, interconnected by FDR InfiniBand. We use NVIDIA CUDA 10.1 and its default profiler for time measurement. We implement \textsc{ClickTrain} based on PyTorch \cite{paszke2019pytorch} using SGD. We evaluate \textsc{ClickTrain} and compare it with state-of-the-art methods on seven well-known CNNs including ResNet18/32/50/101 and VGG11/13/16. Our datasets include CIFAR10/100 \cite{CIFAR} and ImageNet-2012 \cite{ILSVRC}. Note that all accuracy shown in the following evaluation are based on the average of 10 experiments (variance lower than 0.2\%).

Regarding hyperparamter, we initialize the learning rate as 0.1 and set the regularization penalty coefficient as 0.00025. The threshold in stage 1 used for determining when to start stage 2 is set to 0.027. The number of candidate patterns in stage 3 is set to 12, as suggested by Figure~\ref{fig:accuracy_pattern_selection}. We use the bath size (per GPU) of 128 and 64 for CIFAR and ImageNet, respectively.

\subsection{Model Accuracy and Ratio Evaluation}
\begin{table}%[ht]
\scriptsize\sffamily
\centering
\caption{Comparison between \textsc{ClickTrain} (CLK) and PDT-based method \textsc{PruneTrain} (PRT). FLOPs are the saved FLOPs.}
\label{table:result_pattern}
\renewcommand{\arraystretch}{1.0}
\begin{tikzpicture}[y=-1cm]
    \node[rotate=90] at (0.18, -1.75) {\color{B}\fontfamily{ugq}\selectfont CIFAR10};
    \node[rotate=90] at (0.18,  1.60) {\color{B}\fontfamily{ugq}\selectfont CIFAR100};
    \node[rotate=90] at (0.18,  3.65) {\color{B}\fontfamily{ugq}\selectfont \scalebox{0.85}{Image}};
    \node[rotate=90] at (0.38,  3.65) {\color{B}\fontfamily{ugq}\selectfont \scalebox{0.85}{Net}};
    \node[anchor=west] {\begin{adjustbox}{max width=\linewidth}
\begin{tabular}{@{} r r l r r r r @{}}
      \begin{minipage}{6em} \end{minipage}
  &   \fontfamily{ugq}\selectfont \begin{minipage}{3em}\raggedleft PDT\\Method\end{minipage}
  &   \fontfamily{ugq}\selectfont \begin{minipage}{3em}\raggedleft Base.\\Acc.\end{minipage}
  &   \fontfamily{ugq}\selectfont \begin{minipage}{3em}\raggedleft Valid.\\Acc. $\Delta$\end{minipage}
  &   \fontfamily{ugq}\selectfont \begin{minipage}{3.5em}\raggedleft Comp.\\Ratio\end{minipage}
  &   \fontfamily{ugq}\selectfont \begin{minipage}{5em}\raggedleft Train./Inf.\\FLOPs\end{minipage}
  &   \fontfamily{ugq}\selectfont \begin{minipage}{4em}\raggedleft Hard Pr.\\Epoch\end{minipage}
\\[1.5ex]
    \toprule
    % \multicolumn{6}{c}{\textbf{CIFAR-10}} \\ 
    % \cmidrule(l){2-7}
    \multirow{3}{*}{{\hspace{3em}\bfseries ResNet32}} 
  & \textsc{PRT} 
  & 93.6\%
  &  $-$1.8\% 
  & 2.2$\times$ 
  & 53\% / 66\% 
  & \textcolor{gray}{\scshape n/a}\\
    
  & CLK 
  & 93.6\%
  & 0$\pm$0.05\% 
  & 8.6$\times$ 
  & 41.3\% / 85.1\% 
  & 98\\
    
  & \textbf{CLK} 
  & 93.6\%
  & \textbf{0$\pm$0.07}\% 
  & \textbf{10.7$\times$} 
  & \textbf{43.0}\% / \textbf{85.7}\% 
  & 95\\
    \cmidrule(l){2-7}
    \multirow{3}{*}{\textbf{{ResNet50}}} 
  & \textsc{PRT} 
  & 94.2\%
  &  $-$1.1\% 
  & 2.3$\times$ 
  & 50\% / 70\% 
  & \textcolor{gray}{\scshape n/a}\\
    
  & CLK 
  & 94.1\%
  & 0$\pm$0.04\% 
  & 8.5$\times$ 
  & 37.5\% / 74.3\% 
  & 95\\
    
  & \textbf{CLK} 
  & 94.1\%
  & \textbf{ $-$0.2$\pm$0.05\%} 
  & \textbf{10.8$\times$} 
  & \textbf{41.2}\% / \textbf{77.6}\% 
  & 90\\
    \cmidrule(l){2-7}
    \multirow{3}{*}{\textbf{{VGG11}}} 
  & \textsc{PRT} 
  & 92.1\%
  & $-$0.7\% 
  & 8.1$\times$ 
  & 57\% / 65\% 
  & \textcolor{gray}{\scshape n/a}\\
    
  & CLK 
  & 92.1\%
  & $-$0.1$\pm$0.04\% 
  & 8.7$\times$ 
  & 41.2\% / 81.5\% 
  & 96\\
    
  & \textbf{CLK} 
  & 92.1\%
  & \textbf{ $-$0.3$\pm$0.06\%} 
  & \textbf{11.5$\times$} 
  & \textbf{43.9}\% / \textbf{85.3}\% 
  & 94\\
    \cmidrule(l){2-7}
    \multirow{3}{*}{\textbf{{VGG13}}}
  & \textsc{PRT} 
  & 93.9\%
  & $-$0.6\% 
  & 8.0$\times$ 
  & 56\% / 63\% 
  & \textcolor{gray}{\scshape n/a}\\
    
  & CLK 
  & 93.8\%
  & 0$\pm$0.08\% 
  & 8.6$\times$ 
  & 41.3\% / 81.3\% 
  & 95\\
    
  & \textbf{CLK} 
  & 93.8\%
  & \textbf{ $-$0.2$\pm$0.04\%} 
  & \textbf{10.9$\times$} 
  & \textbf{42.5}\% / \textbf{84.9}\% 
  & 96\\
    % \midrule
    % \multicolumn{6}{c}{\textbf{CIFAR-100}} \\
    \midrule
    \multirow{3}{*}{\textbf{{ResNet32}}}
  & \textsc{PRT} 
  & 71.0\%
  &  $-$1.4\% 
  & 2.1$\times$ 
  & 32\% / 46\% 
  & \textcolor{gray}{\scshape n/a}\\
    
  & CLK 
  & 71.0\%
  & 0$\pm$0.05\% 
  & 8.3$\times$ 
  & 41.7\% / 82.9\% 
  & 95\\
    
  & \textbf{CLK} 
  & 71.0\%
  & \textbf{$-$0.2$\pm$0.05}\% 
  & \textbf{10.4$\times$} 
  & \textbf{45.2\%} / \textbf{85.6\%} 
  & 90\\
    \cmidrule(l){2-7}
    \multirow{3}{*}{\textbf{{ResNet50}}}
  & \textsc{PRT} 
  & 73.1\%
  & $-$0.7\% 
  & 1.9$\times$ 
  & 53\% / 69\% 
  & \textcolor{gray}{\scshape n/a}\\
    
  & CLK 
  & 73.1\%
  & 0$\pm$0.04\% 
  & 8.2$\times$ 
  & 36.7\% / 73.6\% 
  & 96\\
    
  & \textbf{CLK} 
  & 73.1\%
  & \textbf{$-$0.2$\pm$0.07}\% 
  & \textbf{9.7$\times$} 
  & \textbf{38.9\%} / \textbf{77.3}\% 
  & 95\\
    \cmidrule(l){2-7}
    \multirow{3}{*}{\textbf{{VGG11}}} 
  & \textsc{PRT} 
  & 70.6\%
  & $-$1.3\% 
  & 3.0$\times$ 
  & 47\% / 57\% 
  & \textcolor{gray}{\scshape n/a}\\
    
  & CLK 
  & 70.6\%
  & 0$\pm$0.1\% 
  & 6.7$\times$ 
  & 40.1\% / 78.6\% 
  & 95\\
    
  & \textbf{CLK} 
  & 70.6\%
  & \textbf{$-$0.2$\pm$0.06}\% 
  & \textbf{8.4$\times$} 
  & \textbf{43.1\%} / \textbf{82.0\%} 
  & 92\\
    \cmidrule(l){2-7}
    \multirow{3}{*}{\textbf{{VGG13}}} 
  & \textsc{PRT} 
  & 74.1\%
  & $-$1.1\% 
  & 2.9$\times$ 
  & 42\% / 52\% 
  & \textcolor{gray}{\scshape n/a}\\
    
  & CLK
  & 74.1\% 
  & $-$0.1$\pm$0.05\% 
  & 7.4$\times$ 
  & 40.5\% / 79.7\% 
  & 95\\
    
  & \textbf{CLK} 
  & 74.1\%
  & \textbf{$-$0.2$\pm$0.08}\% 
  & \textbf{9.2$\times$} 
  & \textbf{41.7\%} / \textbf{83.3\%} 
  & 96\\    
    % \cmidrule(l){2-7}
    % \multicolumn{6}{c}{\textbf{ImageNet}} \\
    % \cmidrule(l){2-7}
    \midrule
        % \multirow{3}{*}{ResNet50} 
  & \textsc{PRT} 
  & 76.2\%
  &  $-$1.9\% 
  & 2.1$\times$ 
  & 40\% / 53\% 
  & \textcolor{gray}{\scshape n/a}\\
    
    \bfseries{ResNet50}
  & \textbf{CLK} 
  & 76.2\%
  &  \textbf{$-$0.6$\pm$0.07\%} 
  & \textbf{4.3$\times$} 
  & \textbf{36.9\%}  / \textbf{66\%} 
  & 40\\  
    \bottomrule        
\end{tabular}
\end{adjustbox}};
\end{tikzpicture}
\end{table}

We first evaluate our proposed \textsc{ClickTrain} on different CNNs and datasets with a fixed number of training epochs, as shown in Table \ref{table:result_pattern}. 
We use $\Delta$ to denote the validation accuracy drop of the pruned model compared to the original baseline model. 
Note that at the end of the regularized training stage, we conduct a hard-pruning step (as shown in Figure~\ref{fig:overall_horizontal}) to eventually zero out the pruned weights which have been regularized to tiny values. Thus, the rest of training process can be accelerated through our compiler-assisted training optimizations.
We train the baseline models to a high accuracy using 190 epochs and 90 epochs for CIFAR10/100 and ImageNet, respectively. 
It is worth noting that all the above baseline accuracies have been widely used in many previous studies \cite{lym2019prunetrain,niu2020patdnn,ma2020image}. 
We thus use the same 190 epochs and 90 epochs for \textsc{ClickTrain} on CIFAR and ImageNet, respectively.
We calculate the compression ratio only considering the convolutional layers in the models, since the convolutional layers in ResNet and VGG models dominate most of the computation overhead in both training and inference processes. 
In particular, the compression ratio is calculated as 
$\frac{\text{the total number of weights}}{\text{the total number of nonzero weights}}$ based on all the convolutional layers.

Table \ref{table:result_pattern} illustrates that for ResNet32 and ResNet50, our \textsc{ClickTrain} can provide more than 10$\times$ compression ratio while maintaining up to 0.2\% accuracy drop on the two CIFAR datasets, compared to the baseline accuracy. 
For VGG11/13, \textsc{ClickTrain} can also provide 8.6$\times$ to 11.5$\times$ compression ratio with up to 0.3\% accuracy drop. 
Since most of the weights in the convolutional layers have been pruned, \textsc{ClickTrain} can save the training FLOPs and inference FLOPs by 36.7\%$\sim$45.2\% and 73.6\%$\sim$85.7\%, respectively, on CIFAR10/100 with the assist of compiler optimizations. To demonstrate the effectiveness on large dataset, we also train ResNet50 on ImageNet using \textsc{ClickTrain}. It provides a compression ratio of 4.3$\times$ and saves the training/inference FLOPs by 36.9\%/66\%. 

Note that the higher the compression ratio, the faster the inference of pruned model. For example, previous studies \cite{niu2020patdnn, rumi2020accelerating} 
proved that for ResNet50 on ImageNet, a compression ratio of about 4.3 (as same as \textsc{ClickTrain}'s) with the pattern-based pruning (adopted by \textsc{ClickTrain}) reduces the model inference latency (batch size 1) by 11$\times$ on the Qualcomm Snapdragon 855 mobile platform (with a Qualcomm Kryo 485 octa-core CPU); in comparison, a previous work \cite{wang2020pruning} illustrates that a compression ratio of about 2.1 (as same as \textsc{Prunetrian}'s) with the structure pruning (adopted by \textsc{Prunetrian}) reduces the inference latency by only 0.72$\times$ on an Intel Xeon E5-2680 v4 CPU. Note that the single-core performance difference between Kryo 485 and Xeon E5-2680 v4 is within 5\% \cite{cpu-comparison}. 

\begin{table}
\scriptsize\sffamily
    \centering
    \caption{Comparison between \textsc{ClickTrain} and PAT-based methods on ImageNet. Well-train costs about 90 epochs.}
    \label{table:result_other}
    \renewcommand{\arraystretch}{1.0}
    \begin{adjustbox}{max width=\linewidth}
\begin{tabular}{@{}r l *{4}{r} @{}}
	&   \fontfamily{ugq}\selectfont \begin{minipage}{3em}\raggedleft PAT\\Method\end{minipage}
	&   \fontfamily{ugq}\selectfont \begin{minipage}{3em}\raggedleft Base.\\Acc.\end{minipage}
	&   \fontfamily{ugq}\selectfont \begin{minipage}{3em}\raggedleft Valid.\\Acc. $\Delta$\end{minipage}
	&   \fontfamily{ugq}\selectfont \begin{minipage}{3em}\raggedleft Comp.\\Ratio\end{minipage}
	&   \fontfamily{ugq}\selectfont \begin{minipage}{4em}\raggedleft Total\\Epochs\end{minipage}
\\[1.5ex]
    \toprule
    \multirow{3}{*}{\textbf{ResNet-18}} 
	& \makebox[2.5em][l]{TAS} \cite{dong2019network} 
	& 70.6\% 
	& $-$1.5\% 
	& 1.5$\times$ 
	& 120 
\\
	& \makebox[2.5em][l]{DCP} \cite{zhuang2018discrimination} 
	& 69.6\% 
	& $-$5.5\% 
	& 3.3$\times$ 
	& well train + 60 
\\
	& \textbf{CLK} 
	& 69.6\% 
	& \textbf{$-$0.9}\% 
	& \textbf{4.1$\times$} 
	& \textbf{90} 
\\
    % \midrule
    \cmidrule(l){2-6}
    \multirow{3}{*}{\textbf{ResNet-50}}
	& \makebox[2.5em][l]{GBN} \cite{you2019gate} 
	& 75.8\% 
	& $-$0.6\% 
	& 2.2$\times$ 
	& well train + 60 
\\
	& \makebox[2.5em][l]{GAL} \cite{lin2019towards} 
	& 76.4\% 
	& $-$7.1\% 
	& 2.5$\times$ 
	& well train + 30 
\\
	& \textbf{CLK} 
	& 76.1\% 
	& \textbf{$-$0.7}\% 
	& \textbf{4.3$\times$} 
	& \textbf{90} 
\\
    \cmidrule(l){2-6}
    % \midrule
    \multirow{2}{*}{\textbf{ResNet-101}} 
	& \makebox[3.8em][l]{RSNLIA} \cite{ye2018rethinking}
	& 75.27\% 
	& $-$2.1\% 
	& 1.9$\times$ 
	& well train + tune 
\\
	& \textbf{CLK} 
	& 76.4\% 
	&\textbf{$-$1.2}\% 
	& \textbf{4.2$\times$} 
	& \textbf{90} 	
\\
    \cmidrule(l){2-6}
    % \midrule
    \multirow{2}{*}{\textbf{VGG-16}} 
	& \makebox[2.5em][l]{NeST} \cite{dai2019nest} 
	& 71.6\% 
	& $-$2.3\% 
	& 6.5$\times$ 
	& N/A 
\\
	& \textbf{CLK} 
	& 73.1\% 
	&\textbf{$-$0.8}\% 
	& \textbf{6.6$\times$} 
	& \textbf{90} 
\\
    \bottomrule        
\end{tabular} 
\end{adjustbox}
\end{table}

We also evaluate the impact of when to start the hard pruning on the validation accuracy and FLOPs. We observe that the earlier the patterns and unimportant kernels are determined, the earlier the models can be hard pruned by \textsc{ClickTrain}, thus more training FLOPs can be saved; however, earlier hard pruning causes a significant accuracy loss. The epoch of hard pruning shown in Table \ref{table:result_pattern} makes a good tradeoff between accuracy and training FLOPs.

\begin{figure}[b]
	\centering
	\begin{subfigure}{\linewidth}
    \centering
	\includegraphics[width=0.98\linewidth]{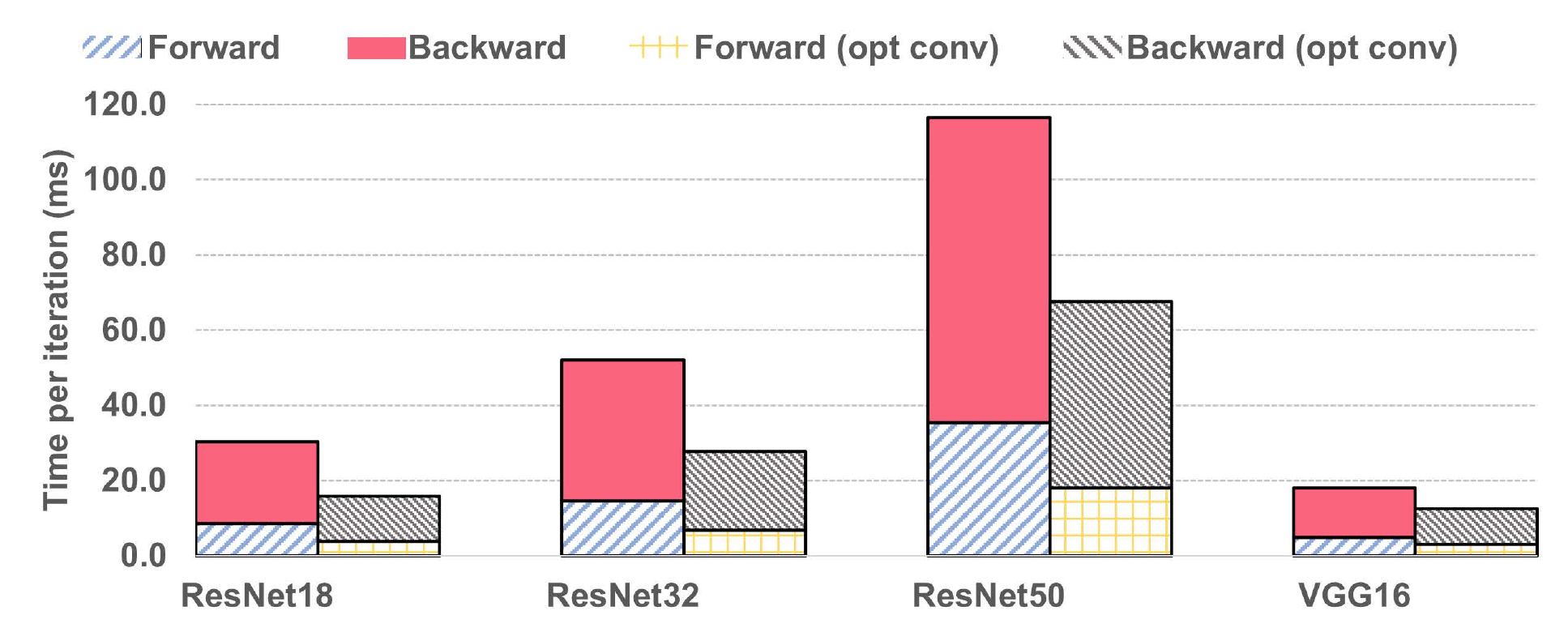}
    \caption{CIFAR}
	\label{fig:forward_bp_time_cifar}
    \end{subfigure}

    \begin{subfigure}{\linewidth}
    \centering
	\includegraphics[width=0.98\linewidth]{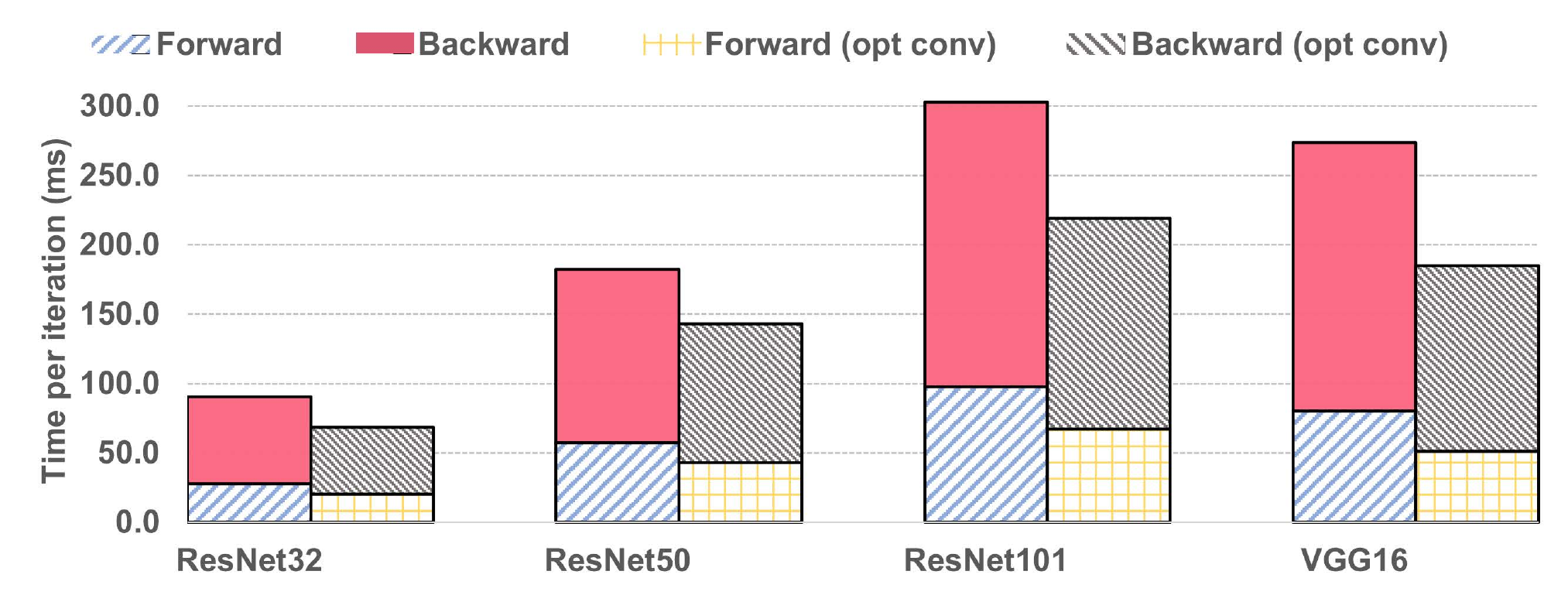}
    \caption{ImageNet}
	\label{fig:forward_bp_time_imagenet}
    \end{subfigure}
    \caption{Average forward and backward time per iteration on CIFAR and ImageNet using a single GPU. ``opt conv'' means solution with our optimized SpMM for sparse convolution.}
	\label{fig:train_cifar_1}
\end{figure}

{\bf Comparison with PDT-based Method.}
We then compare our \textsc{ClickTrain} with the state-of-the-art PDT-based method \textsc{PruneTrain}, as shown in Table \ref{table:result_pattern}.
It illustrates that \textsc{ClickTrain} can precisely control the accuracy drop within -0.3\% for all the tested models on CIFAR10/100, but the accuracy drop of \textsc{PruneTrain} is over -1.0\% for most of the tested CNN models. 
Moreover, \textsc{ClickTrain} can significantly improve the compression ratio with similar or even higher accuracy, compared to \textsc{PruneTrain}.
For example, \textsc{PruneTrain} can only provide a compression ratio less than 3$\times$ with more than 1.0\% accuracy drop for ResNet50 on CIFAR10, whereas \textsc{ClickTrain} achieves 10.8$\times$ compression ratio with only 0.2\% accuracy drop, leading to 4.7$\times$ higher compression ratio. 
For VGG13 on CIFAR100, \textsc{ClickTrain} achieves 9.2$\times$ compression ratio with only 0.2\% accuracy drop, which significantly outperforms \textsc{PruneTrain}'s 2.9$\times$ compression ratio with 1.1\% accuracy drop. 
For ImageNet, \textsc{ClickTrain} reduces the accuracy drop by 1.3\% and improves the compression ratio by 2.1$\times$ over \textsc{PruneTrain}.%, compared to \textsc{PruneTrain}.

\begin{figure}[t]
	\centering
	\begin{subfigure}{\linewidth}
    \centering
	\includegraphics[width=0.98\linewidth]{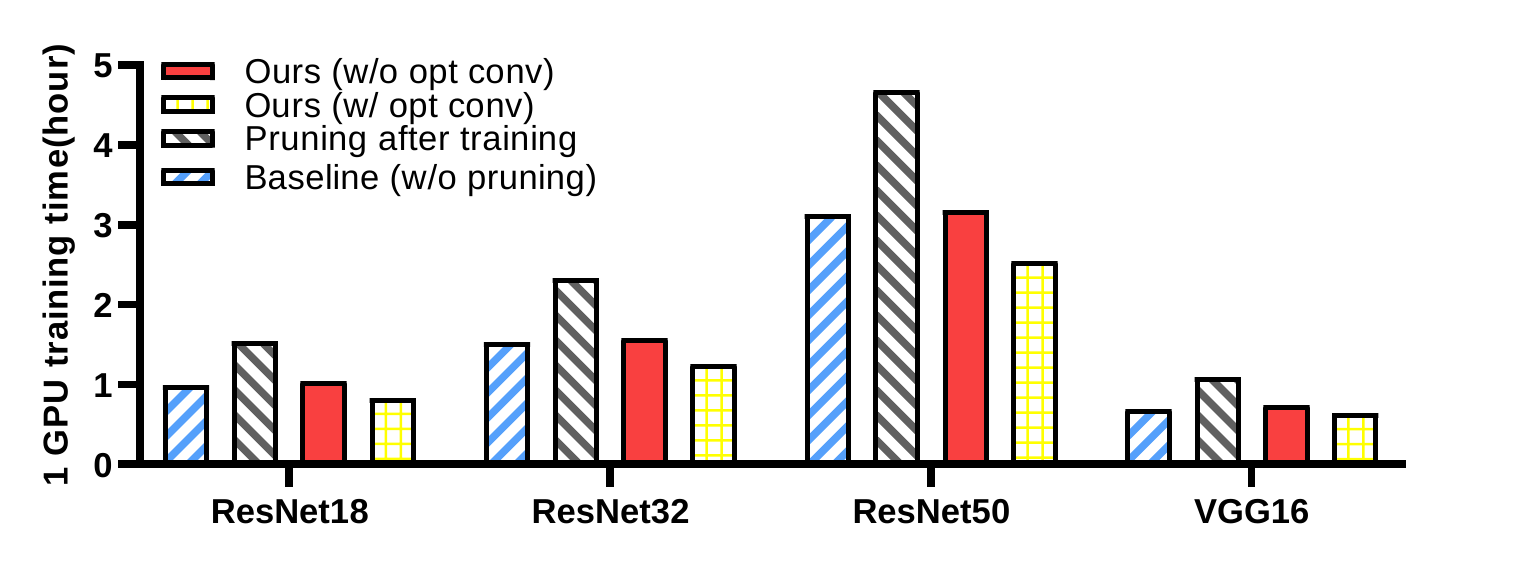}
    \caption{CIFAR10}
	\label{fig:train_cifar_1}
    \end{subfigure}

    \begin{subfigure}{\linewidth}
    \centering
	\includegraphics[width=0.98\linewidth]{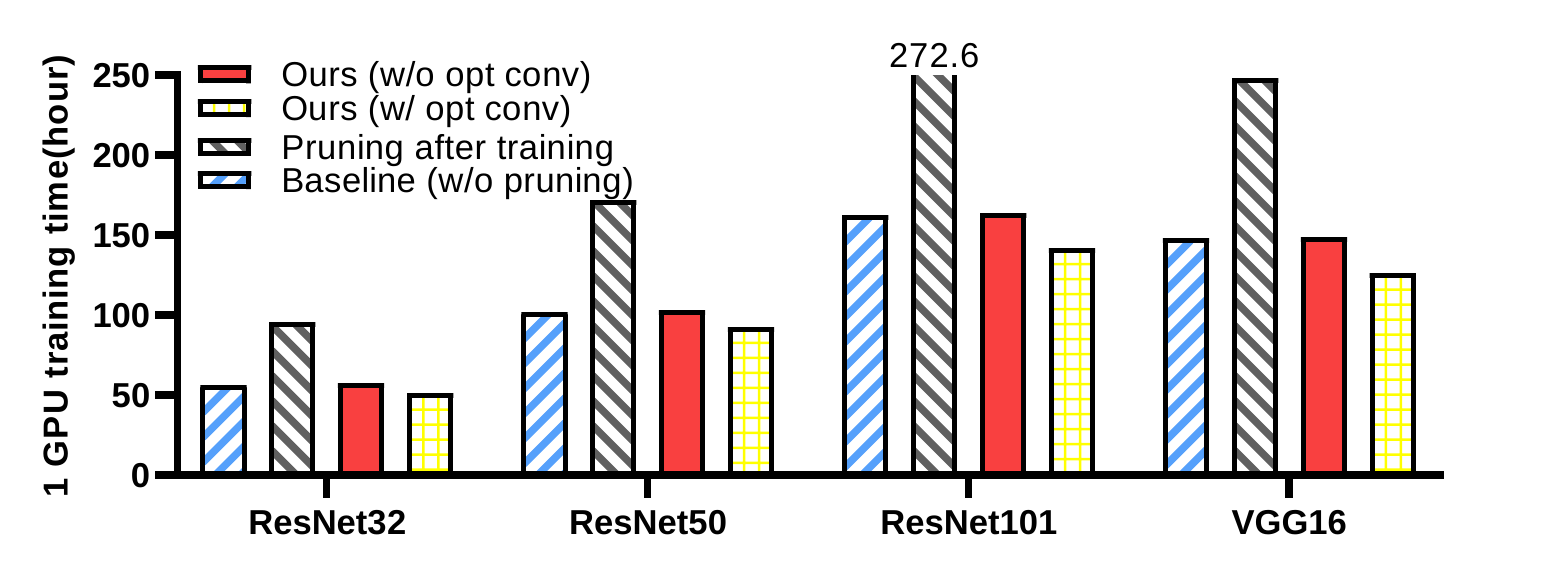}
    \caption{ImageNet}
    \vspace{-2mm}
	\label{fig:train_image_1}
    \end{subfigure}
    \caption{Total training time on CIFAR and ImageNet (single-GPU).}
	\label{fig:train_cifar_1}
\end{figure}

{\bf Comparison with PAT-based Methods.}
A typical procedure of model pruning is removing the redundant weights based on well-trained networks and then fine tuning the slashed networks. Thus, we finally compare our \textsc{ClickTrain} with state-of-the-art PAT-based methods. As illustrated in Table \ref{table:result_other}, \textsc{ClickTrain} save up to about 67\% computation time while only sacrificing up to 1.2\% accuracy, compared with other PAT-based methods.
In addition, \textsc{ClickTrain} also achieves a much higher compression ratio for efficient inference. Thus, we conclude that accurately estimating the weight importance during training makes our PDT-based solution feasible to significantly save the total training epochs.

\subsection{Single-GPU Performance Evaluation}
Then, we evaluate the single-GPU performance gain by our PDT-based algorithm and optimized SpMM of \textsc{ClickTrain}.

{\bf Forward and Backward Acceleration.}
We first evaluate the forward and backward time per iteration on a single GPU when applying our optimized SpMM using different CNN models with CIFAR and ImageNet. The sparsity (i.e., compression ratio) for each model can be found in Table~\ref{table:result_other}.

Figure~\ref{fig:forward_bp_time_cifar} shows that for forward phase on CIFAR, \textsc{ClickTrain} achieves the speedups of 2.2$\times$, 2.1$\times$, and 1.9$\times$, and 1.6$\times$ on ResNet18, ResNet32, ResNet50, and VGG16, respectively. 
For backward phase, \textsc{ClickTrain} achieves the speedups of 1.8$\times$, 1.7$\times$, 1.6$\times$, and 1.3$\times$ on ResNet18, ResNet32, ResNet50, and VGG16, respectively.
Figure~\ref{fig:forward_bp_time_imagenet} shows that for forward phase on ImageNet, \textsc{ClickTrain} achieves the speedups of 1.4$\times$, 1.3$\times$, 1.5$\times$, and 1.6$\times$ on ResNet32, ResNet50, ResNet101, and VGG16, respectively. 
For backward phase on imagenet, \textsc{ClickTrain} achieves the speedups of 1.3$\times$, 1.25$\times$, 1.35$\times$, and 1.5$\times$ on ResNet32, ResNet50, ResNet101, and VGG16, respectively.

{\bf Overall Training Acceleration.}
We then evaluate the training acceleration of \textsc{ClickTrain} with the four CNNs on CIFAR10 and ImageNet, as shown in Figure~\ref{fig:train_cifar_1}.
Compared with the baseline training (i.e., training without pruning), \textsc{ClickTrain} saves 0.16, 0.29, 0.59, and 0.15 hours on ResNet18/32/50 and VGG16, respectively, on CIFAR10. 
When training on ImageNet, \textsc{ClickTrain} saves 5.1, 9.4, 20.4, and 19.7 hours on ResNet32/50/101 and VGG16, respectively.

\subsection{Multi-GPU Performance Evaluation}
\begin{figure}
	\centering
	\includegraphics[width=0.9\linewidth]{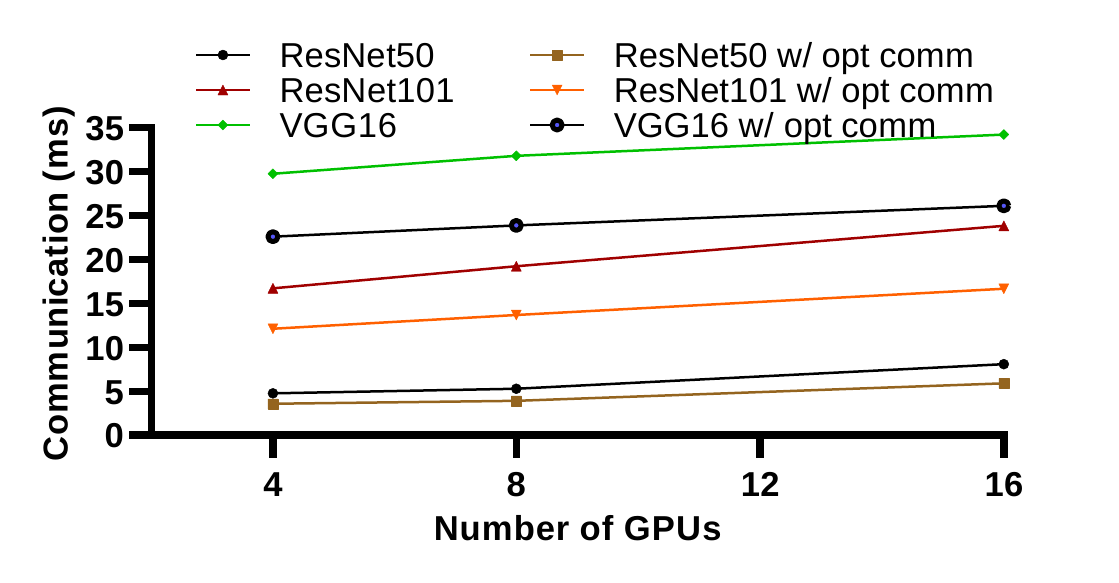}
	\vspace{-2mm}
    \caption{Communication time of baseline training and \textsc{ClickTrain} with different numbers of GPUs. ``opt comm'' means solution with our optimized communication.}
	\label{fig:comm_delay}
\end{figure}

{\bf Communication Time.} 
Next, we evaluate our optimized communication time using multiple GPUs on ResNet50, ResNet101 and VGG16 with ImageNet, as shown in Figure~\ref{fig:comm_delay}. For ResNet50, \textsc{ClickTrain} can save 23.9\%, 26.3\%, and 28.4\% communication time per iteration using 4, 8, and 16 GPUs, respectively. For ResNet101, \textsc{ClickTrain} can save 27.5\%, 29.7\%, and 31.2\% communication time per iteration using 4, 8, and 16 GPUs, respectively. For VGG16, \textsc{ClickTrain} can save 21.2\%, 24.3\%, and 24.9\% communication time per iteration using 4, 8, and 16 GPUs, respectively. 
We notice that although PyTorch's optimization of communication and computation overlap offsets our optimization for communication overhead to some extent, the performance gain will increase when increasing the scale (i.e., number of nodes and GPUs). 

\begin{figure}
	\centering
	\includegraphics[width=\linewidth]{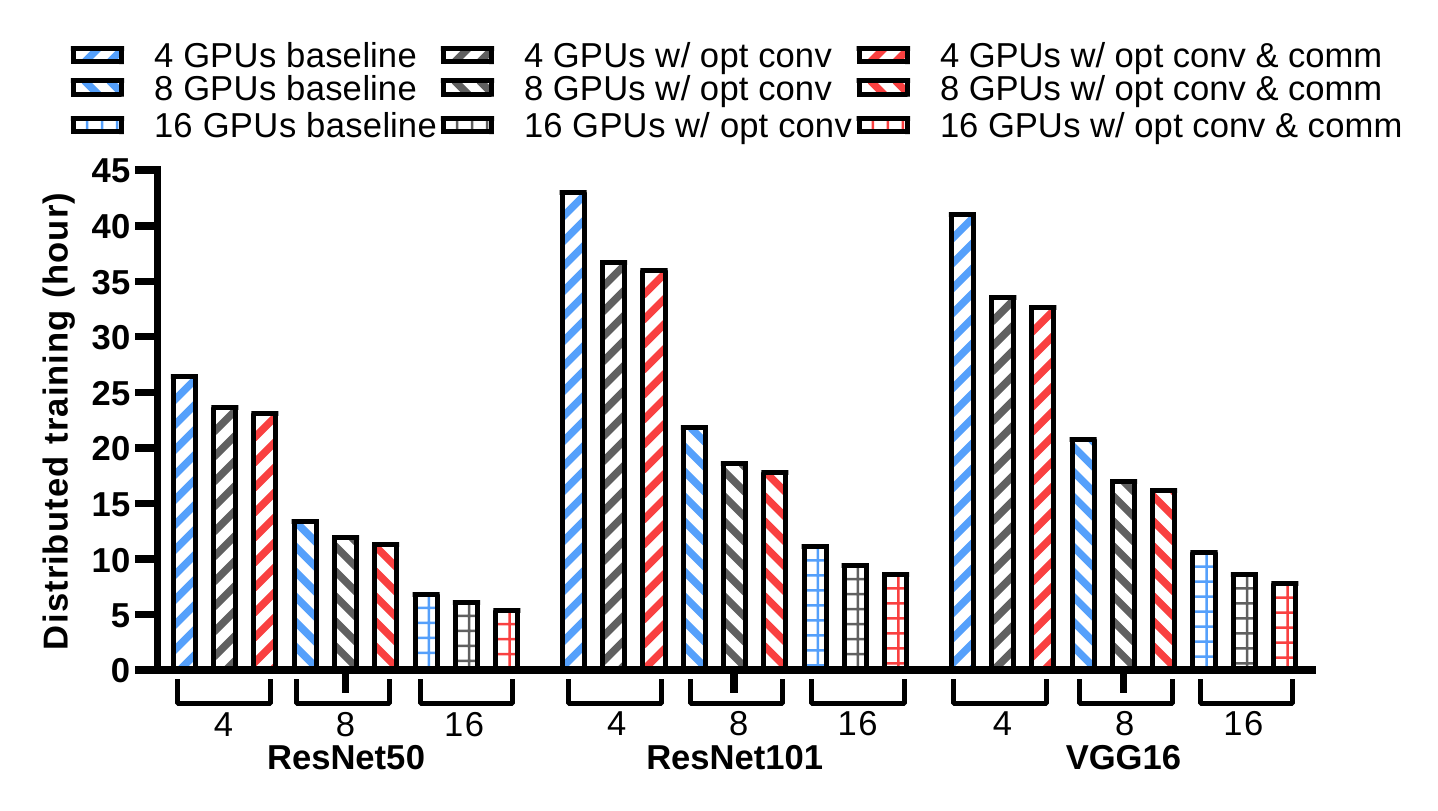}
	\vspace{-4mm}
    \caption{Total time of baseline training and \textsc{ClickTrain}.}
	\label{fig:total_time}
\end{figure}

{\bf Total Training Time.}
Furthermore, we evaluate the overall training time on different CNN models using ImageNet with batch size 64 per GPU. Figure~\ref{fig:total_time} shows that compared with the baseline training, for on ResNet50, \textsc{ClickTrain} saves 3.2 hours, 2.1 hours, and 1.4 hours using 4, 8, and 16 GPUs, respectively; for ResNet101, \textsc{ClickTrain} saves 7.1 hours, 4 hours, and 2.5 hours using 4, 8, and 16 GPUs, respectively; for on VGG16, \textsc{ClickTrain} saves 6.3 hours, 4.6 hours, and 2.7 hours using 4, 8, and 16 GPUs, respectively. Note that unlike the baseline training, \textit{\textsc{ClickTrain} can generate ready-to-deploy models without further tuning such as pruning.}
Moreover, compared with the baseline PAT-based approach (i.e., pattern-based pruning after training), \textsc{ClickTrain} reduces the training time by 1.67$\times$/1.96$\times$/2.13$\times$, 1.86$\times$/1.99$\times$/2.18$\times$, and 1.98$\times$/2.15$\times$/2.25$\times$ using 4, 8, and 16 GPUs, respectively, on ResNet50/ResNet101/VGG16.

\begin{figure}
	\centering
	\includegraphics[width=1.0\linewidth]{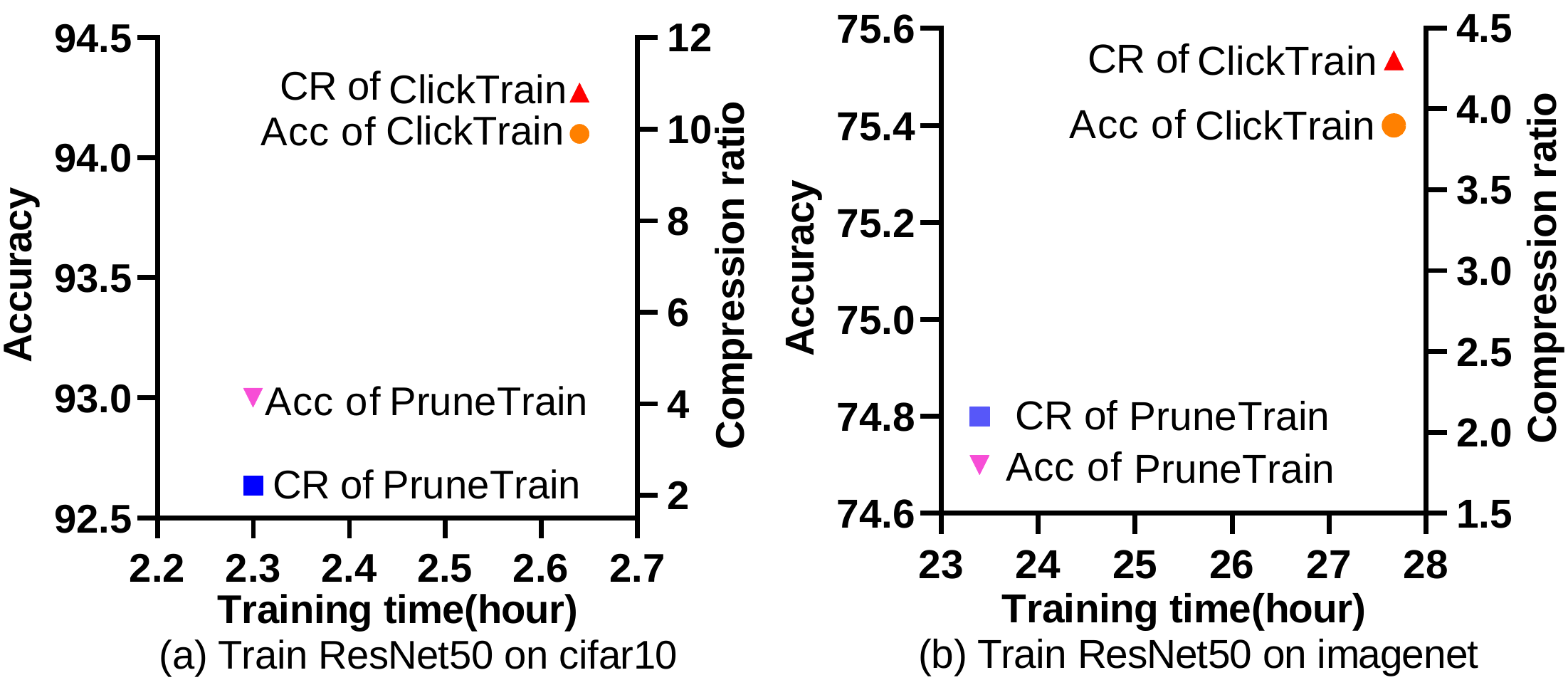}
	\vspace{-4mm}
    \caption{Comparison of \textsc{PruneTrain} and \textsc{ClickTrain}.}
	\label{fig:guide_train}
\end{figure}

{\bf Comparison with PruneTrain.}
Finally, Figure~\ref{fig:guide_train}a shows that \textsc{ClickTrain} achieves a compression ratio of 10.8$\times$ with only 0.2\% accuracy drop on ResNet50 with CIFAR10, whereas \textsc{PruneTrain} only has 2.3$\times$ compression ratio but with a significant accuracy drop of 1.1\%, even it saves 16.7\% end-to-end time over \textsc{ClickTrain}.
Figure~\ref{fig:guide_train}b shows that \textsc{ClickTrain} has 2.1$\times$ higher compression ratio than \textsc{PruneTrain} with a notable accuracy improvement of 1.3\% on ResNet50 with ImageNet but only slightly longer (i.e., 18.4\%) end-to-end time. 
Note that 1.3\% on ImageNet and 0.9\% on CIFAR10 are significant accuracy improvements considering the limited training time. 
Thus, aggressively preserving the original architecture is critical for designing PDT-based approaches.

\section{Related Work}
\label{sec:related}

\textsc{PruneTrain} \cite{lym2019prunetrain} is a state-of-the-art approach to accelerate the DNN training from scratch while pruning it. 
The work observed that when pruning with group-lasso regularization, once a group of model weights are penalized close to zero, their magnitudes are typically impossible to recover during the rest of the training process.
Based on this observation, \textsc{PruneTrain} periodically removes the small weights and reconfigure the network architecture and hence can gradually reduce the training during training toward both high compression ratio and accuracy.  
In addition, \textsc{PruneTrain} also proposes to dynamically increase the mini-batch to further increase the training performance. 
However, as we discussed in Section \ref{sec:motivation}, \textsc{PruneTrain} aggressively change the original network architecture, causing a significant unrecoverable accuracy loss. 

The lottery ticket hypothesis \cite{frankle2018lottery} proves that subnetworks can match the test accuracy of original network after training for at most the same number of iterations. Those subnetworks are termed as winning tickets, which can be directly trained efficiently. However, it is challenging to determine the architecture of subnetwork. 

The Early-Bird (EB) tickets work \cite{you2019drawing} claims that winning tickets can be identified at a very early training stage using aggressively low-cost training algorithms. Even though EB also adopts the strategy that firstly trains a few epochs and then prunes the network, it chooses channel pruning, which will change the network architecture and lead to a significant accuracy drop.

\section{Conclusion}
\label{sec:con}

In this paper, we propose \textsc{ClickTrain} by using dynamic fine-grained pattern-based pruning. It has both algorithm-level and system-level optimizations with four stages.  i) accurate weight importance estimation to select the pattern, ii) dynamic pattern generation and finalization, iii) regularized training for fine-tuning with an enhanced group-lasso, and iv) compiler-assisted optimized training. 
%We show that \textsc{ClickTrain} significantly reduces the end-to-end time while still achieving high accuracy and compression ratio.
Our experimental results on seven DNNs and three datasets demonstrate that \textsc{ClickTrain} can reduce the cost of the state-of-the-art PAT-based method by up to 2.3$\times$ with comparable accuracy and compression ratio. Compared with the state-of-the-art training acceleration approach, \textsc{ClickTrain} can improve the pruned accuracy by up to 1.8\% and the compression ratio by up to 4.9$\times$ on the tested CNNs and datasets, with comparable training time. We plan to extend \textsc{ClickTrain} to more types of DNNs in the future.

\newpage
\section*{Acknowledgments}
This material is based upon work supported by National Science Foundation under Grant No. OAC-2034169, OAC-2042084, CCF-1937500, CNS-1909172, IIS-1850546, IIS-2008973, CNS-1951974. The work was also partially supported by Jeffress Trust Awards in Interdisciplinary Research, Facebook faculty award and Australian Research Council Discovery Project DP210101984. The authors would like to thank the Texas Advanced Computing Center (TACC) for providing access to the Frontera supercomputer.
\bibliographystyle{ACM-Reference-Format}
\bibliography{references}

\end{document}